\documentclass[a4paper]{article}

\usepackage[latin1]{inputenc}
\usepackage[T1]{fontenc}

\usepackage{fullpage}         
\usepackage[english]{babel}
\usepackage{amsthm,amssymb,amsfonts,amsmath,color,graphicx,float,url,bbm,multirow,caption,subfig,placeins,xspace}
\usepackage[authoryear,round,sort]{natbib}

\addto\extrasenglish{
  
}

\numberwithin{equation}{section}

\newtheorem{Ass}{Assumption}[section]
\newtheorem{rem}{Remark}[section]
\newtheorem{defi}{Definition}[section]
\newtheorem{lem}{Lemma}[section]

\newtheorem{pro}{Proposition}[section]
\newtheorem{theo}{Theorem}[section]

\newcommand{\C}{\mathcal{C}}
\newcommand{\T}{\mathcal{T}}

\renewcommand{\c}{\mathbf{c}}
\renewcommand{\r}{\mathbf{r}}

\renewcommand{\P}{\mathbb{P}}
\newcommand{\e}{\varepsilon}
\newcommand{\f}{\mathbf{f}}

\def\e{\varepsilon}
\def\E{\mathbb{E}}
\def\I{\mathbf{1}}
\def\R{\mathbb{R}}

\def\1{\mathbf{1}}

\title{Aggregation using input-output trade-off}
\author{Aur\'elie Fischer \& Mathilde Mougeot\\\\Laboratoire de Probabilités et Modèles Aléatoires\\Université Paris Diderot\\75013 Paris, France}

\definecolor{vert1}{rgb}{0.0, 0.5, 0.0}

\begin{document}
\maketitle

\begin{abstract}
In this paper, we introduce a new learning strategy based on a seminal idea of Mojirsheibani (1999, 2000, 2002a, 2002b), who proposed a smart method for combining several classifiers, relying on a consensus notion.
In many aggregation methods, the prediction for a new observation $x$ is computed by building a linear or convex combination
over a collection of basic estimators $r_1(x), \ldots, r_m(x)$ previously calibrated using a training data set.
Mojirsheibani proposes to compute the prediction associated to a new observation by combining selected outputs of the training examples.
The output of a training example is  selected if  some kind of consensus  is observed:
  the predictions computed for the  training example with the different machines 
have to be ``similar'' to the prediction for the new observation.
This approach has been recently extended   to the context of regression in \cite{Cobra}.

In the original scheme, the agreement condition is actually required to hold for all individual estimators, which appears inadequate if there is one bad initial estimator. In practice, a few disagreements are allowed ; for establishing the theoretical results, the proportion of  estimators satisfying the  condition is required to tend to 1.

In this paper, we propose an alternative procedure, mixing the previous consensus ideas on the predictions with  the Euclidean distance computed between entries.
This may be seen as an alternative approach allowing to reduce the effect of a possibly bad estimator in the initial list,
using a constraint on the inputs.

We prove the consistency of our strategy in classification and in regression. We also provide some numerical experiments on simulated and real data
to illustrate the benefits of this new aggregation method. On the whole, our practical study shows that our method may perform much better than the original combination technique, and, in particular, exhibit  far less variance.
We also show on simulated examples that this procedure mixing inputs and outputs is still robust to high dimensional inputs.

\end{abstract}

\bigskip

\emph{Keywords : Classification, regression estimation,  aggregation, nonlinearity,
consistency.}\\

\emph{AMS classification : 62G05, 62G08, 62H30}

\section{Introduction}

Given the growing number of available statistical estimation strategies, trying to combine several procedures in a smart way is a very natural idea, and so, an abundant literature on aggregation of estimators for different types of statistical models has sprung up in recent years. A widespread aggregation method consists in building a linear or a convex combination of a bunch of initial estimators (see, for instance, \cite{Cat04},
\cite{JN00}, \cite{Nem00}, \cite{Yang00,
	Yang01, Yang04}, \cite{GKKW}, \cite{W03}, \cite{Aud}, \cite{BTW06,BTW07,BTW07a}, and \cite{DalTsy}. Observe that the model selection approach, which aims at selecting the best estimator in the list, is also linked to the same goal (see, for example, the monograph by \cite{MassStF}).

In this paper, we  focus on the question of combining estimators for  two kinds of statistical procedures, pertaining to the supervised learning framework:  classification, which consists in assigning to an observation a  label, out of a finite number of candidates, and regression, whose aim is to associate real outputs to entries.

Beside the usual linear aggregation and model selection  methods, a quite different  point of view has been introduced by \cite{Majid} for classification.
The idea consists in an original combination method, which is  non linear in the initial estimators and is based on a consensus concept.
 More specifically, in this combining scheme, an observation is considered to be reliable for contributing to the classification of a new query point if all initial classifiers predict the same label for both points. Then, the label for the query point is estimated thanks to a majority vote among the labels of the observations which have been retained this way.
Note that more regular versions, based on smoothing kernels, have also been developed (\cite{Majid1}). A numerical comparison study of several combining schemes is available in \cite{Majid3}, and recently, a variant of the method has been proposed in \cite{BalaMoj}.

This strategy has just been adapted in the regression framework by \cite{Cobra}. In this  context, an observation is used in the combination step if  all the initial estimators predict a similar value for the observation and the new point: the difference between  both predictions is required to be less than some prespecified threshold. Then, the new prediction by the combined estimator is the average of the outputs corresponding to the selected entries. Note  that the functional data  framework  has also been considered, by \cite{Frai}.

In the classification case as well as in the regression one,  when the initial  list contains a  consistent estimator, it can be shown that the combined estimator inherits this consistency property. Let us mention that the techniques of proof, and consequently, the assumptions made in both situations, are quite different. For instance, the number of initial estimators is expected to tend to infinity with the sample size  in \cite{Majid1} whereas it is fixed in \cite{Cobra}.

It is worth pointing out that, in both contexts described above, the condition for an observation to be reliable is in principle required  to be satisfied  for all estimators. In a further paper, \cite{Majid2}  notes that this rule may seem too restrictive and proposes to allow a few disagreements (typically, a single one). The resulting classifier is still consistent provided that the number of initial classifiers  keeps tending to infinity after removing those with disagreement.  Similarly, in \cite{Cobra}, this unanimity constraint is relaxed  in practice by demanding that the distance condition for keeping an observation is true at least for a certain proportion $\alpha$ of the estimators (for example, $\alpha=80\%$). The theoretical results remain true provided that $\alpha$ tends to 1.

Here, our purpose is to investigate a new approach, based on distances between observations, which also aims at reducing the effect of a possibly bad initial estimator. Roughly, choosing a kernel point of view, we will propose a combined estimator with weights constructed by mixing distances between entries with distances between predictions coming from the individual estimators.
Our motivation for introducing such a strategy is the intuition that taking advantage of the efficiency of the consensus idea of \cite{Majid} and \cite{Cobra} without for all that forgetting the information related to the proximity between entries shall help improving the prediction, especially in the presence of an initial estimator that does not perform very well.

Our modified rule will be shown to be consistent under general assumptions. In particular, the combined estimator may perfectly be consistent even if  the list of initial estimators does not contain any consistent estimator. We also conduct numerical experiments, both on simulated and real data, which demonstrate the benefits of our strategy, with respect to the original combining method and the individual estimators.

The paper is organized as follows. The new combined estimator is defined in Section \ref{section:df}. Then, the main theoretical results are stated in Section \ref{section:main}.
Section \ref{section:simu} is devoted to numerical experiments with simulated and real examples. A few perspectives are presented in a brief  conclusive paragraph, Section \ref{section:ccl}. For the sake of clarity, proofs  are postponed to Section \ref{section:proof}.

\section{Notation and  definition of the estimator}\label{section:df}
Let $(X,Y)$ denote a random pair  taking its values in $\R^d\times \mathcal Y$. The variable $X$ has distribution $\mu$. We are interested in two different  situations: $\mathcal Y=\{0,1\}$, which corresponds to the binary classification problem, and $\mathcal Y=[0,1]$, that is bounded regression. Let $\eta$ stand for the regression function  $\eta(x)=\E[Y|X=x]$. Note that $\eta(x)=\P(Y=1|X=x)$ in the classification context.

Let $\psi^\star$ denote the Bayes classifier, given by
$$\psi^\star(x)=\begin{cases}1 &\mbox{if }\eta(x)>1/2\\0&\mbox{otherwise}.
\end{cases}$$ It is well-known that $\psi^\star$ minimizes over all possible classifiers $\psi$ the missclassification error $L(\psi)=\P(\psi(X)\neq Y)$.

Throughout the document, we assume that we are given a sample $\mathcal D_n=\{(X_1,Y_1)\dots,(X_n,Y_n)\}$ of the random pair $(X,Y)$.

Our goal in regression is to estimate the function $\eta$ using $\mathcal D_n$. In classification, we aim at building a classifier based on $\mathcal D_n$ whose error mimics the Bayes classifier error.

For every $k\geq 1,$ if $u\in \R^k$, $\|u\|_k$ will denote the Euclidean norm of the vector $u$, and $B_k(x,r)=\{y\in \R^k,\|x-y\|_k\leq r\}$ the closed ball centered at $x\in\R^k$, with radius $r>0$. To simplify notation, as there is no ambiguity, we will most of the time drop the index $k$ and simply write $\|u\|$ and $B(x,r)$.

Let $K:\R^{d+p}\mapsto \R_+$ be a kernel, that is a nonnegative and monotone decreasing function along rays starting from
the origin.
The next assumption will be made on the kernel $K$ (see \cite{DGL}).

\begin{Ass}

We suppose that the kernel $K$ is regular, that is, there exist $c>0$ and $\rho>0$ such that
\begin{itemize}

\item For all $z$, $K(z)\geq c\I_{B_{d+p}(0,\rho)}(z)$.
\item $\int\sup_{t\in B_{d+p}(z,\rho)}K(t)dz<\infty.$

\end{itemize}
\end{Ass}

We propose to combine the predictions  of $p$ initial estimators, denoted by $c_1,\dots,c_p$ in the classification context or $r_1,\dots,r_p$ in the regression context. Sometimes, the notation  $f_1,\dots,f_p$, embracing  both cases, will  be employed. For $x\in\R^d$, let $\c(x)=(c_1(x),\dots,c_p(x))$,  $\r(x)=(r_1(x),\dots,r_p(x))$ and $\f(x)=(f_1(x),\dots,f_p(x))$.
For ease of exposition, we will assume throughout  that $c_1,\dots,c_p$ and $r_1,\dots,r_p$ do not depend on the sample $\mathcal D_n$.

\begin{rem}
Using a simple sample-splitting device, our results extend to the case where the individual estimators depend on  the sample. More specifically, we may split $\mathcal D_n$ into two independent parts $\mathcal D_{n_1}$ and $\mathcal D_{n_2}$,  assuming  that $f_1,\dots,f_p$ are based on $\mathcal D_{n_1}$ alone, and using $\mathcal D_{n_2}$ for the combination step.
\end{rem}

The definition of our combined estimator is first introduced in the regression framework.

Let the function $g:\R^d\times \R^p\to\R_+$ be such that $g(v_1,v_2)=K(v)$, where $v=(v_1,v_2)\in\R^{d+p}$.

\begin{defi}

Suppose that we are given a set of initial regression estimators $r_1,\dots,r_p$. 
 The regression  combined estimator $\T_n$ is  defined by 
\begin{align*}\T_n(x)&=
\frac{\sum_{i=1}^n Y_i K(\frac{Z_i^r-z^r}{\alpha})}{\sum_{i=1}^n  K(\frac{Z_i^r-z^r}{\alpha})}\\&=\frac{\sum_{i=1}^n Y_i g\left(\frac{X_i-x}{\alpha},
\frac{\r(X_i)-\r(x)}{\beta}\right)}{\sum_{i=1}^ng\left(\frac{X_i-x}{\alpha},
\frac{\r(X_i)-\r(x)}{\beta}\right)},\end{align*}

where $Z_i^r=(X_{i1},\dots, X_{id},\frac\alpha\beta r_1(X_i),\dots,\frac\alpha\beta r_p(X_i))$, $i=1,\dots,n$,  $z^r=(x_{1},\dots, x_{d},\frac\alpha\beta r_1(x)\dots,\frac\alpha\beta r_p(x))$.
\end{defi}

To lighten the equations, we will sometimes use the notation  $x\mapsto K_\alpha( x)$ to mean the function $x\mapsto K(\frac x{\alpha})$ and $(v_1,v_2)\mapsto g_{\alpha,\beta}(v_1,v_2)$ for $(v_1,v_2)\mapsto g(\frac{v_1}\alpha,\frac{v_2}\beta)$.

We introduce the notation $\T_n^\star$, for  the quantity defined by $$\frac{\sum_{i=1}^n Y_i g\left(\frac{X_i-x}{\alpha},
\frac{\f(X_i)-\f(x)}{\beta}\right)}{n\E\left[g\left(\frac{X-x}{\alpha},
\frac{\f(X)-\f(x)}{\beta}\right)\right]}.$$

We now state the definition of the estimator in the context of classification.
\begin{defi} Suppose that we are given a set of initial classifiers $c_1,\dots,c_p$. 
The combined classifier $\C_n$ be defined by
\begin{align*}\C_n(x)&=\begin{cases}
0 &\mbox{if } \frac{\sum_{i=1}^n Y_i K_\alpha(Z_i^c-z^c)}{n\E [ K_\alpha(Z^c-z^c)]}\leq \frac{\sum_{i=1}^n (1-Y_i) K_\alpha(Z_i^c-z^c)}{n\E[  K_\alpha(Z^c-z^c)]}\\
1 &\mbox{otherwise}\end{cases}
\\&=\begin{cases}
0 &\mbox{if } \frac{\sum_{i=1}^n Y_i g_{\alpha,\beta}(X_i-x,\c(X_i)-\c(x))}{n\E[  g_{\alpha,\beta}(X-x,\c(X)-\c(x))]}\leq \frac{\sum_{i=1}^n (1-Y_i) g_{\alpha,\beta}(X_i-x,\c(X_i)-\c(x))}{n\E [ g_{\alpha,\beta}(X-x,\c(X)-\c(x))]}\\
1 &\mbox{otherwise},
\end{cases}
\end{align*}
where $Z_i^c=(X_{i1},\dots, X_{id},\frac\alpha\beta c_1(X_i),\dots,\frac\alpha\beta c_p(X_i))$, $i=1,\dots,n$, and $z^c=(x_{1},\dots, x_{d},\frac\alpha\beta c_1(x)\dots,\frac\alpha\beta c_p(x))$.
\end{defi}

\begin{rem}The random vectors $Z_1^r,\dots,Z_n^r$ form an i.i.d. sequence, and so is the sequence  $Z_1^c,\dots,Z_n^c$.

\end{rem}
Here, the exponent $r$ or $c$ will be dropped whenever there is no need to specify the classification or regression context.

\begin{rem}
In the particular case where $K$ can be expressed in function of the squared Euclidean norm, i.e. we have, for $x\in\R^{d+p}$, $K(x)=H(\|x\|^2)$, then the quantity $K(\frac{Z-z}{\alpha})=g\left(\frac{X-x}{\alpha},
\frac{\r(X)-\r(x)}{\beta}\right)$ takes the somewhat more explicit form $H\left( \big\|\frac{X-x}{\alpha}\big\|^2
+\big\|\frac{\r(X)-\r(x)}{\beta}\big\|^2\right)$. This is the case for a Gaussian kernel for instance.
\end{rem}

\section{Main results}\label{section:main}

We are now in a position to state the main results of the paper.

\begin{theo}[Regression case]\label{theo:reg} If $\alpha\to 0$ and $n\alpha^d\beta^p\to \infty$ as $n\to\infty$, then, for every $\e>0$, there exists $n_0$ such that for $n\geq n_0$, the following exponential bound holds:
$$\P\left(\int|\eta(x)-\T_n(x)|\mu(dx)>\e\right)\leq 2\exp\left(-\frac{n\e^2}{32R^2}\right), $$
where  $R\geq 0$ is a constant depending on $K$ and $d+p$.
\end{theo}
Let $L^\star$ denote the Bayes error and $L_n$ the missclassification error of $\C_n$.

\begin{theo}[Classification case]\label{theo:class} If $\alpha\to 0$ and $n\alpha^d\beta^p\to \infty$ as $n\to\infty$, then, for every $\e>0$, there exists $n_0$ such that for $n\geq n_0$, the following exponential bound holds:
$$\P(L_n-L^\star>\e)\leq 2\exp\left(-\frac{n\e^2}{32R^2}\right), $$
where  $R\geq 0$ is a constant depending on $K$ and $d+p$.
\end{theo}


\bigskip

These results may be seen as ``combining'' versions of the strong consistency  results for kernel regression and the kernel classification rule, described in \cite{DK} (see also the books of \cite{DGL} and \cite{GKKW}).

For Theorem \ref{theo:reg} (regression context), let us write
\begin{equation}\label{eq:dec1}\int |\eta(x)-\T_n(x)|\mu(dx)\leq \int|\eta(x)-\T_n^\star(x)|\mu(dx)+\int |\T_n^\star(x)-\T_n(x)|\mu(dx).\end{equation}

Thus, the result will be obtained by replacing, on the one hand, $\T_n$ by $\T_n^\star$,  and by controlling, on the other hand, the error due to the difference between the two terms. This  is done respectively in Lemma \ref{lem:etaTn*} and \ref{lem:TnTn*} below.

\begin{lem}\label{lem:etaTn*}If $\alpha\to 0$ and $n\alpha^d\beta^p\to \infty$ as $n\to\infty$, then there exists $R\geq 0$, depending on $K$ and $d+p$, such that for every $\e>0$, if  $n$ is large enough, 
$$\P\left(\int|\eta(x)-\T_n^\star(x)|\mu(dx)>\e/2\right)\leq \exp\left(-\frac{n\e^2}{32R^2}\right).$$
\end{lem}


The next lemma is devoted to the control of the difference between $\T_n$ and $\T_n^\star$.

 \begin{lem}\label{lem:TnTn*}
If $\alpha\to 0$ and $n\alpha^d\beta^p\to \infty$ as $n\to\infty$, then there exists $R\geq 0$, depending on $K$ and $d+p$, such that  for every $\e>0$, if  $n$ is large enough, $$\P\left(\int |\T_n^\star(x)-\T_n(x)|\mu(dx)>\e/2\right)\leq \exp\left(-\frac{n\e^2}{8R^2}\right).$$
\end{lem}

For Theorem \ref{theo:class}  (classification context), let us note that Theorem 2.3 in \cite{DGL} may be adapted to obtain  the next lemma. 
For the sake of completeness, it is proved in section \ref {section:proof}.

\begin{lem}\label{lem:regclas}
The following upper bound holds:
 $$L_n-L^\star\leq \int\left|1-\eta(x)-\frac{\sum_{i=1}^n  K_\alpha(Z_i-z)}{n\E  [K_\alpha(Z-z)]}+\T_n^\star(x)\right|\mu(dx)+\int|\eta(x)-\T_n^\star(x)|\mu(dx).$$

\end{lem}

We deduce from this upper bound that 
\begin{multline}\label{eq:declass}\P(L_n-L^\star>\e)\leq \P\left(\int\left|1-\eta(x)-\frac{\sum_{i=1}^n  K_\alpha(Z_i-z)}{n\E  [K_\alpha(Z-z)]}+\T_n^\star(x)\right|\mu(dx)+\int|\eta(x)-\T_n^\star(x)|\mu(dx)>\e\right)
\\\leq \P\left(\int\left|1-\eta(x)-\frac{\sum_{i=1}^n  K_\alpha(Z_i-z)}{n\E  [K_\alpha(Z-z)]}+\T_n^\star(x)\right|\mu(dx)>\e/2\right)\\
+\P\left(\int|\eta(x)-\T_n^\star(x)|\mu(dx)>\e/2\right)
.\end{multline}
Consequently, Theorem \ref{theo:class}  (classification context) follows from  Lemma \ref{lem:etaTn*},  applied to $Y$ and to $1-Y$.

\medskip

To control the quantity $\int|\eta(x)-\T_n^\star(x)|\mu(dx)$ in order to prove Lemma \ref{lem:etaTn*}, we may use the following decomposition, for $x\in\R^d$:
\begin{equation}\label{eq:dec2}
|\eta(x)-\T_n^\star(x)|=\E [|\eta(x)-\T_n^\star(x)|]+(|\eta(x)-\T_n^\star(x)|-\E [|\eta(x)-\T_n^\star(x)|]).
\end{equation}
The term $\int\E [|\eta(x)-\T_n^\star(x)|]\mu(dx)$ will be studied first and  then McDiarmid's inequality will be employed to handle the deviation $\int|\eta(x)-\T_n^\star(x)|\mu(dx)-\int\E [|\eta(x)-\T_n^\star(x)|]\mu(dx)$.

Let $\e '>0$ and $f:\R^d\to [0,1]$ (recall that $Y$ equals 0 or 1 in the classification case and we consider bounded regression with $Y\in[0,1]$) be a continuous function with compact support such that \begin{equation}\int |\eta(x)-f(x)|\mu(dx)<\e ' .\label{eq:fccs}\end{equation}

The following inequality holds:
\begin{multline}\label{eq:decoup} 
\E [|\eta(x)-\T^\star_n(x)|]\leq |\eta(x)-f(x)|+\left|f(x)-\frac{\E\left[ f(X)g\left(\frac{X-x}{\alpha},
\frac{\f(X)-\f(x)}{\beta}\right)\right]}{\E \left[g\left(\frac{X-x}{\alpha},
\frac{\f(X)-\f(x)}{\beta}\right)\right]}\right|
\\+\left|\frac{\E\left[ f(X)g\left(\frac{X-x}{\alpha},
\frac{\f(X)-\f(x)}{\beta}\right)\right]}{\E \left[g\left(\frac{X-x}{\alpha},
\frac{\f(X)-\f(x)}{\beta}\right)\right]}-\E [\T^\star_n(x)]\right|+\E|\E [\T_n^\star(x)]-\T_n^\star(x)|.
\end{multline}
We will derive an  upper bound for the integral against $\mu$ of each term. The next result is proved in Section \ref{section:proof}.

\begin{pro}\label{prop:dec}
For a function $f$ as defined in \eqref{eq:fccs}, the following statements hold. 
\begin{enumerate}
\item If $\alpha\to 0$ as $n\to \infty$, then $$\displaystyle\limsup_{n\to\infty}\int \left|f(x)-\frac{\E\left[ f(X)g\left(\frac{X-x}{\alpha},
\frac{\f(X)-\f(x)}{\beta}\right)\right]}{\E \left[g\left(\frac{X-x}{\alpha},
\frac{\f(X)-\f(x)}{\beta}\right)\right]}\right|\mu(dx)\leq \e '.$$
\item We have $\displaystyle \int \left|\frac{\E\left[ f(X)g\left(\frac{X-x}{\alpha},
\frac{\f(X)-\f(x)}{\beta}\right)\right]}{\E \left[g\left(\frac{X-x}{\alpha},
\frac{\f(X)-\f(x)}{\beta}\right)\right]}-\E [\T^\star_n(x)]\right|\mu(dx)\leq R\e '.$
\item Let us write $$\int \E[|\E [\T_n^\star(x)]-\T_n^\star(x)|]\mu(dx)=\int_{[-\ell,\ell]^d} \E[|\E [\T^\star_n(x)]-\T^\star_n(x)|]\mu(dx)+\int_{([-\ell,\ell]^d)^c} \E[|\E [\T^\star_n(x)]-\T^\star_n(x)|]\mu(dx).$$ Then, $$\int_{[-\ell,\ell]^d} \E[|\E [\T^\star_n(x)]-\T^\star_n(x)|]\mu(dx)\leq C(n\alpha^d\beta^p)^{-1/2},$$ for some constant $C\geq 0$ depending on $\ell$, the kernel $K$ and the dimensions $d$ and $p$. Moreover, if $\alpha\to 0$ as $n\to \infty$, then, for $\ell$ and $n$ large enough,  $$\int_{([-\ell,\ell]^d)^c} \E[|\E [\T^\star_n(x)]-\T^\star_n(x)|]\mu(dx)\leq \e'.$$
\end{enumerate}

\end{pro}


The proof of this proposition rests upon the following important result, which extends the covering lemma of \cite{DK} to our context.

\begin{lem}[Covering lemma]\label{lem:cov}
\begin{enumerate}
\item There exists $R\geq 0$, depending on $K$ and $d+p$, such that  $$
\sup_u\int  \frac{g\left(\frac{u-x}{\alpha},
\frac{\f(u)-\f(x)}{\beta}\right)}
{\E \left[g\left(\frac{X-x}{\alpha},
\frac{\f(X)-\f(x)}{\beta}\right)\right]}\mu(dx)
\leq R<+\infty.
$$
\item  $\forall \delta, \e$, there exists $\alpha_0 $ such that $$\sup_{u,\alpha\leq\alpha_0}\int  \frac{g\left(\frac{u-x}{\alpha},
\frac{\f(u)-\f(x)}{\beta}\right)\I_{\{\|x-u\|\geq \delta\}}}
{\E\left[g\left(\frac{X-x}{\alpha},
\frac{\f(X)-\f(x)}{\beta}\right)\right]}\mu(dx)\leq \e .$$
\end{enumerate}
\end{lem}




\section{Numerical Experiments}\label{section:simu}

The section presents numerical experiments to illustrate the benefits of using the new combining approach.
The classification  case is illustrated with numerical simulations and
the regression  case with real operational data recorded from two  applications:
modeling of the electrical power consumption of an air compressor
and  modeling of  the electricity production of  different wind turbines in a wind farm.

Two aggregation strategies are run, the strategy by \cite{Majid} or \cite{Cobra}, which only combines output predictions, and our strategy, which combines the input distance and output  predictions. In the sequel, the original approach is called \textit{Cobra}, from the name of the R package implementing the method  \citep{PackCobra}, and our version is called \textit{MixCobra}.

\subsection{Classification}

\begin{figure}[h!]
	\begin{center}
		\begin{tabular}{ccc}
			\includegraphics[width=4cm, height=4cm]{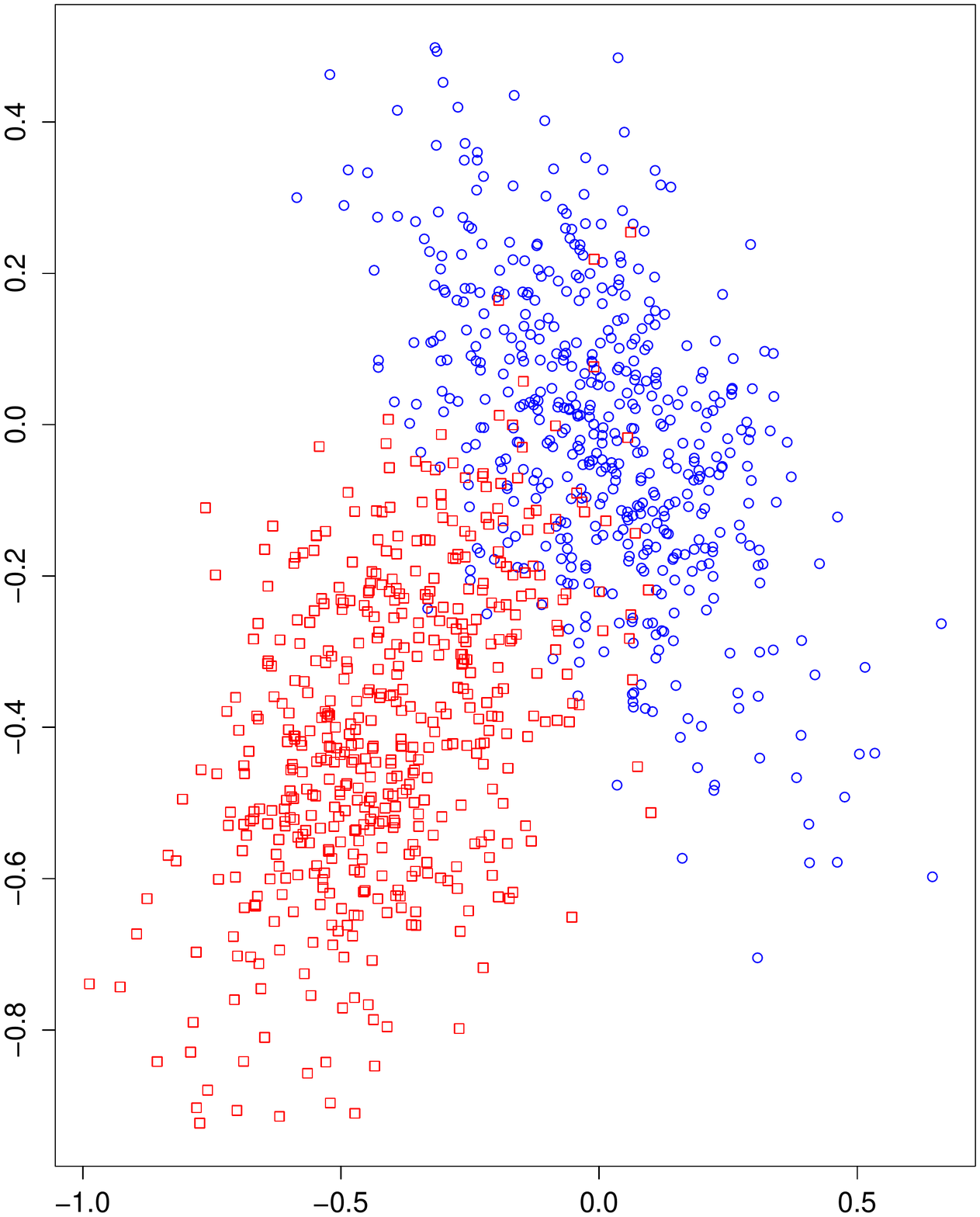}&
			\includegraphics[width=4cm, height=4cm]{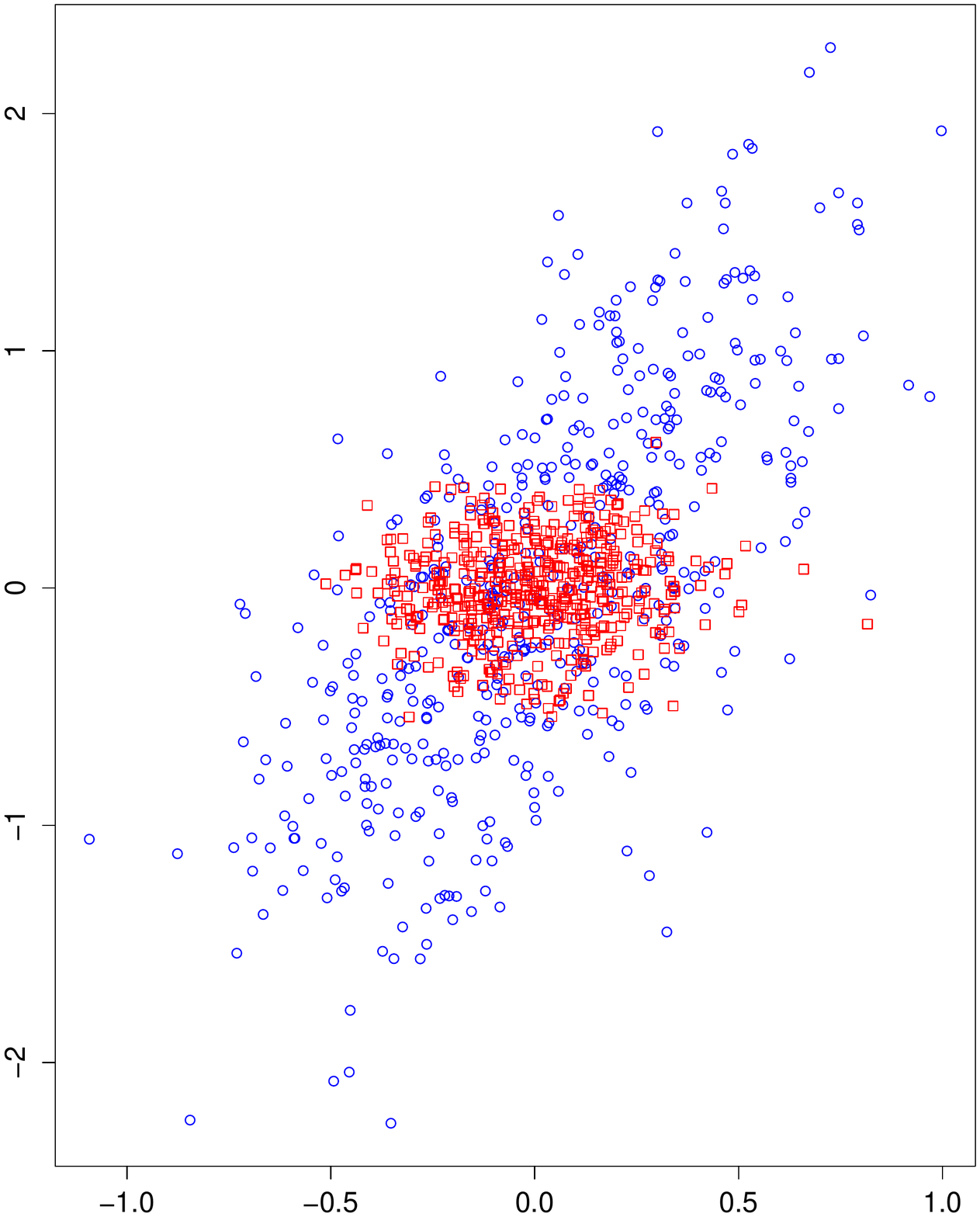}&
			\includegraphics[width=4cm, height=4cm]{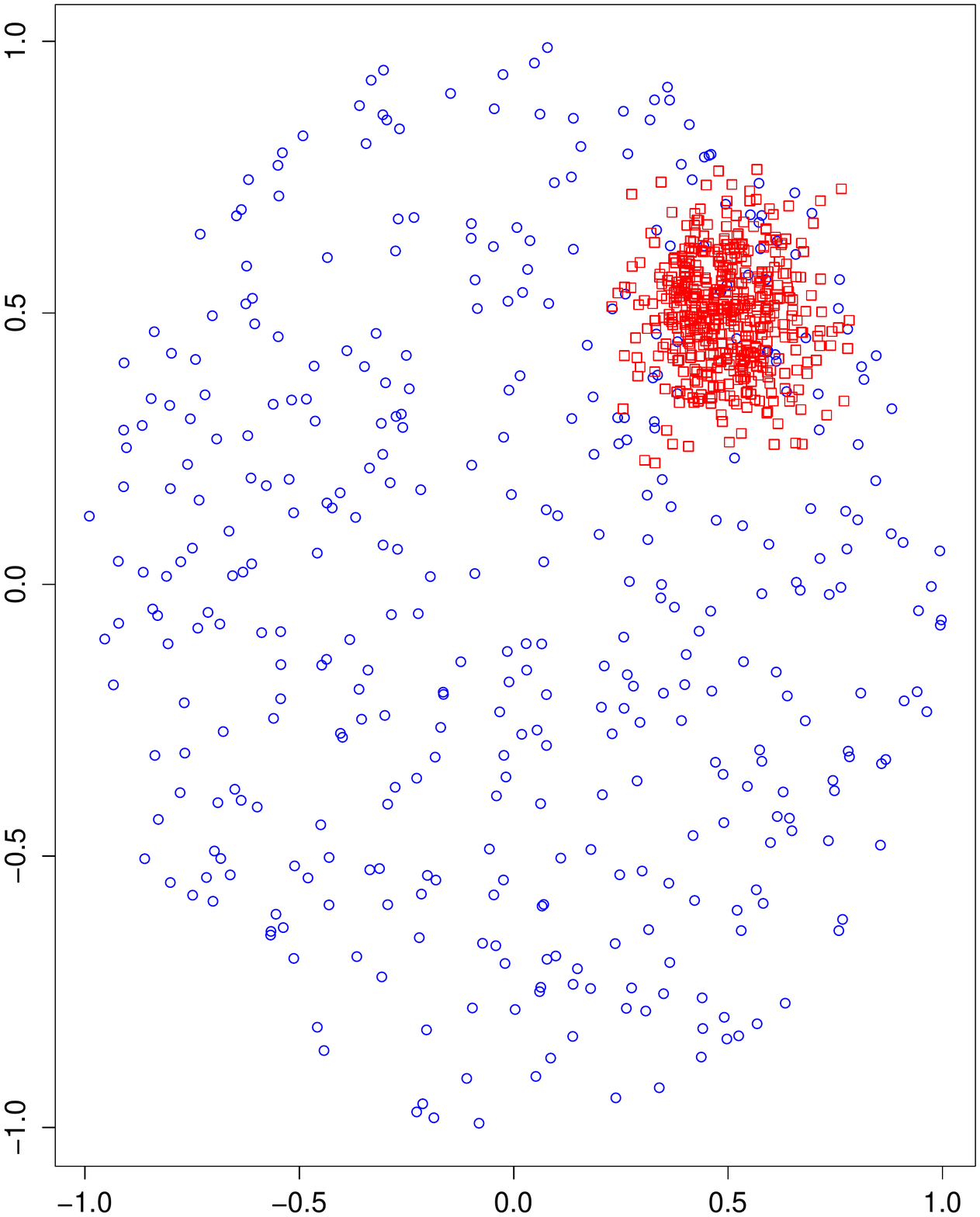}\\
			\includegraphics[width=4cm, height=4cm]{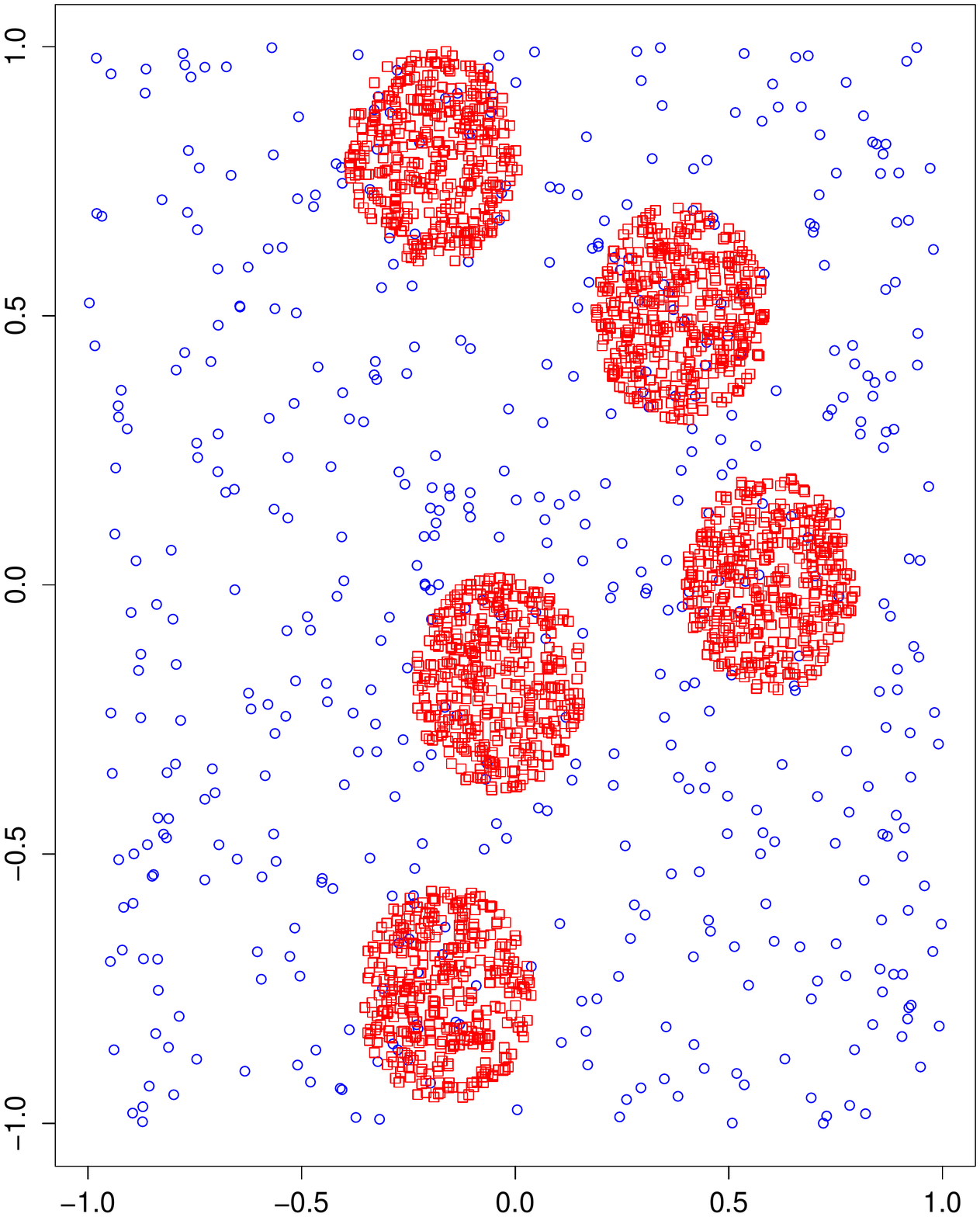}&
			\includegraphics[width=4cm, height=4cm]{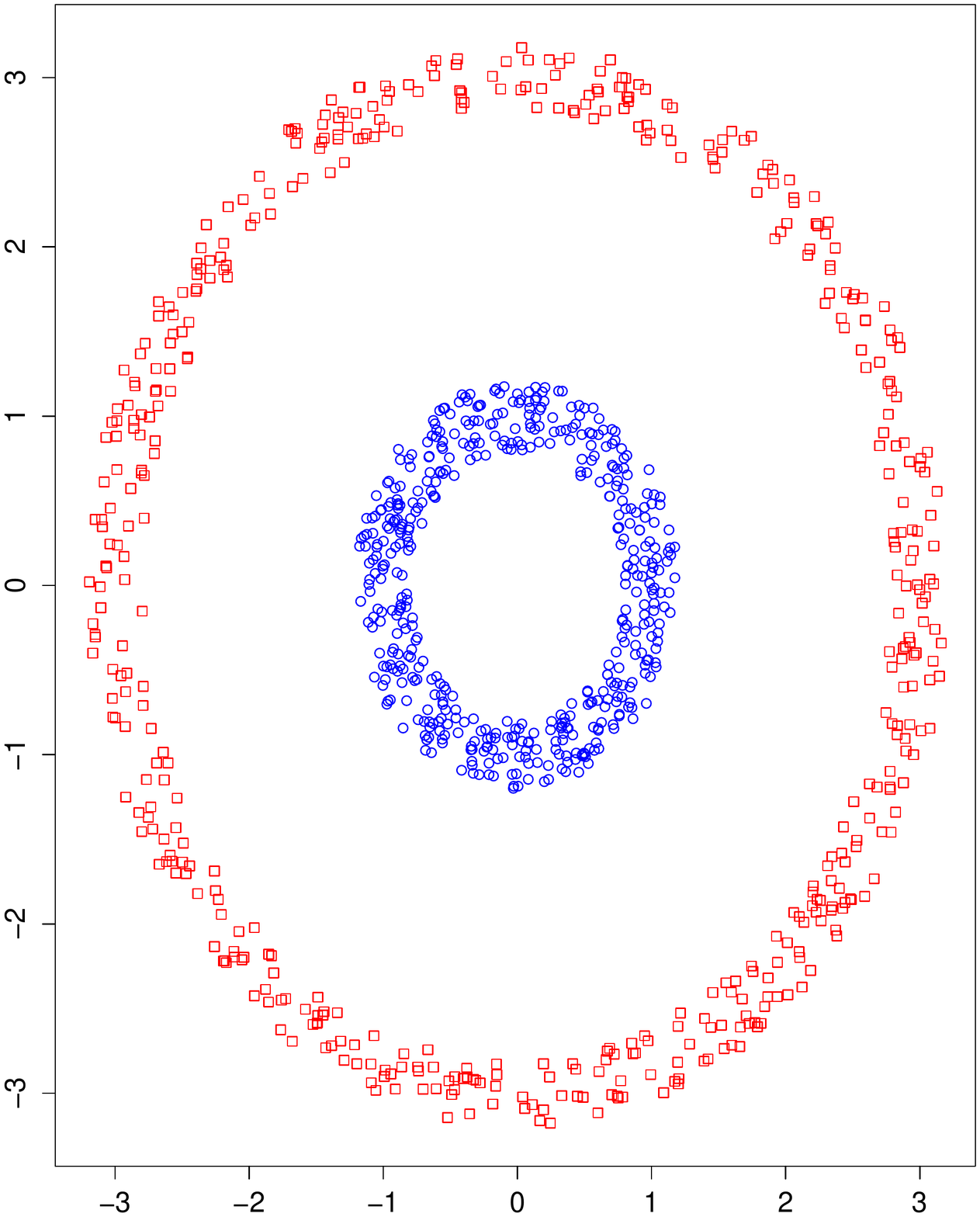}&
			\includegraphics[width=4cm, height=4cm]{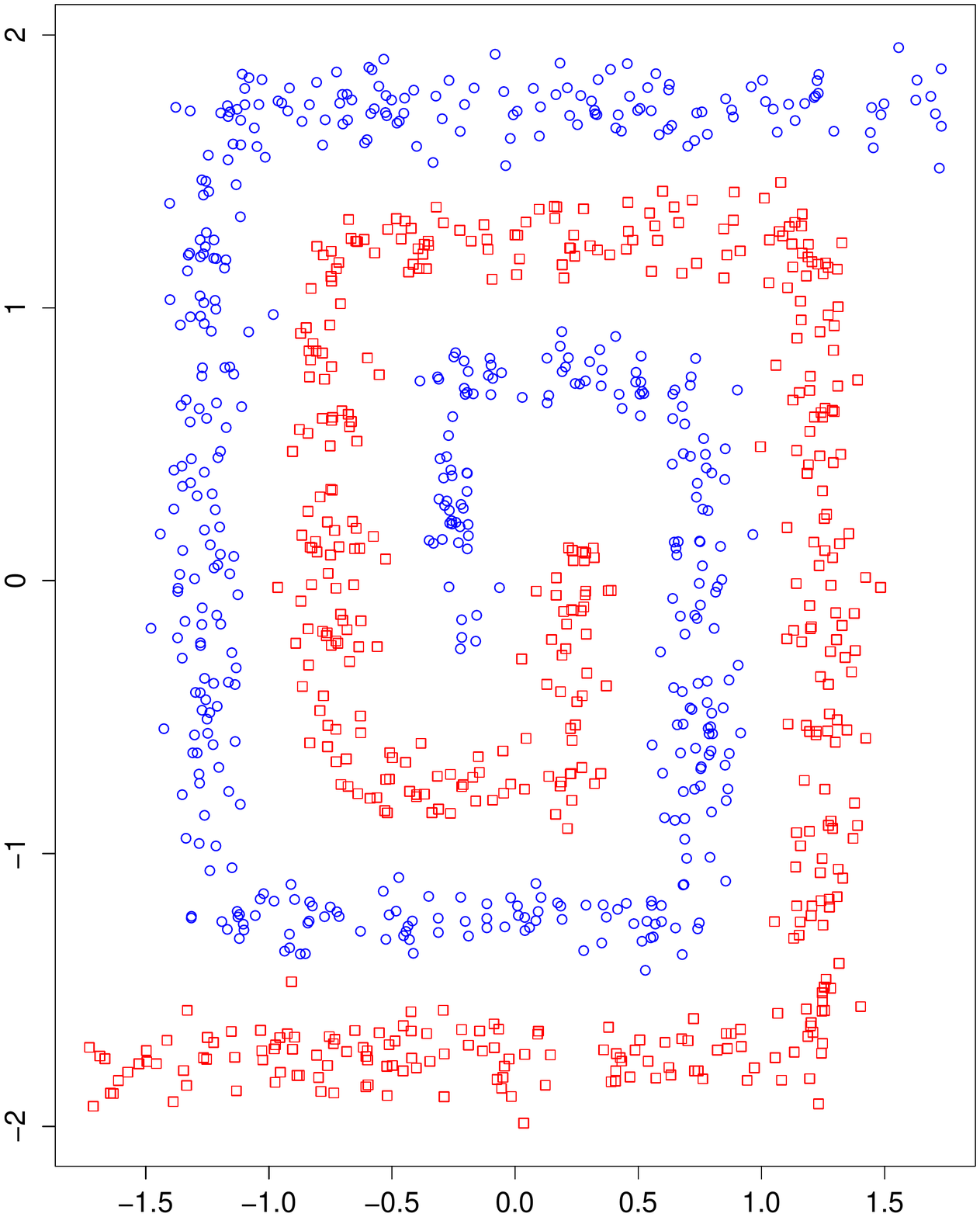}\\
		\end{tabular}
		\caption{Simulated data for classification. From left to right, from up to bottom: ``gauss'', ``comete'', ``nuclear'',
			``spot'', ``circles'' and ``spirals'' examples.
			\label{fig:classif}}
	\end{center}  
\end{figure}

Figure \ref{fig:classif} presents the 6  examples simulated to illustrate the classification performances.
For each example, a data set of $N$  observations is simulated, split into two classes (with half of the observations in each class).
All the figures are depicted with $N=1000$ observations, to get graphics reflecting more accurately    the distribution; 
however,  to be more realistic with respect to available data in most applications,
only $N=200$ observations (100 per class) are used  to evaluate the performances.
The first four graphs present mixtures of uniform or Gaussian distributions.
The two last graphs represent respectively two concentric circles and two interlocked spirals (see annex for details on simulated data).

Parametric and non parametric machines are here introduced for classification.
For parametric methods, linear discriminant analysis (lda) and  logistic regression (logit) are used.
For nonparametric methods, $k$-nearest neighbors (knn) (with $k=5$), support vector machines (svm),
classification and regression trees (cart) and 
random forest (rf) are introduced. In this classification framework, the \textit{Cobra} aggregation procedure is performed using all the available machines,
meaning that the consensus is required for all machines.
For each simulated example, the performances of each machine are computed using
$K=100$ repetitions. For each repetition, $N=200$ observations are generated:
75\% randomly chosen observations are used to calibrate the model and the remaining 25\% to compute
the test performances.

One first interesting question is to analyze whenever a specific machine  always shows the best performance over all the  repetitions.
During an off-line study, if a practitioner observes that a given machine performs always the best, then the choice of the  model to use 
is easy to state  and an aggregation procedure relying on all machines may not be useful in this particularly case.
Table \ref{table:classifrank} shows, for each machine, the number of runs  where this  machine provides the best model (i.e. has the best performance).
For each simulated example (each column), the sum of the rows corresponding to the different methods equals the total number of repetitions ($K=100$).
We observe that, except for the ``circles'' example for which the knn machine outperforms all the others, almost every machine wins at 
least once over all the runs.
For all simulated examples except the ``circles'' example, an aggregation procedure may be interesting.

\begin{table}[ht]
	\centering
	\begin{tabular}{lrrrrrr}
		\hline
		& gauss & comete & nuclear & spot & circles & spirals \\ 
		\hline
		lda & 75 & 9 & 35 & 4 & 0 & 2 \\ 
		logit & 13 & 0 & 13 & 2 & 0 & 0 \\ 
		knn & 5 & 46 & 27 & 44 & 100 & 76 \\ 
		svm & 5 & 20 & 13 & 16 & 0 & 9 \\ 
		cart & 0 & 18 & 8 & 19 & 0 & 2 \\ 
		bag & 1 & 5 & 4 & 12 & 0 & 11 \\ 
		rf & 1 & 2 & 0 & 3 & 0 & 0 \\ 
		\hline
	\end{tabular}
	\caption{Number of runs  where every machine provides the best model for the simulated 
		classification examples (smallest test error).
		\label{table:classifrank}}
\end{table}

Table \ref{table:classif} presents the average  performances for all the classification examples.
As expected for the ``gauss'' example  corresponding to the well separated mixture of two Gaussian distributions,
both parametric and non parametric machines  perform well; otherwise, the nonparametric
methods perform in general better.

\begin{table}[ht]
	\centering
	\begin{tabular}{lllllll}
		\hline
		& gauss & comete & nuclear & spot & circles & spirals \\ 
		\hline
		lda &  7.0 (8.7) & 51.0 (15.6) & 12.9 (10.6) & 17.6 (7.2) & 52.5 (16.5) & 34.3 (15.8) \\ 
		logit &  7.5 (8.2) & 51.1 (15.6) &  9.6 (9.0) & 17.8 (7.3) & 52.4 (16.3) & 34.4 (16.0) \\ 
		knn &  7.0 (8.1) & 25.0 (14.5) &  6.0 (7.1) &  9.2 (6.6) &  0.0 (0.0) &  4.1 (6.4) \\ 
		svm &  8.7 (9.0) & 25.9 (14.6) &  5.9 (8.0) & 10.3 (6.7) &  1.8 (4.1) & 12.2 (10.8) \\ 
		cart &  7.2 (8.3) & 23.0 (14.8) &  4.0 (6.0) &  7.2 (5.8) &  0.0 (0.0) & 17.8 (13.7) \\ 
		bag &  8.7 (8.8) & 28.3 (15.8) &  5.1 (7.5) &  7.5 (5.5) &  0.5 (2.2) &  3.0 (5.4) \\ 
		rf &  8.9 (9.1) & 27.5 (15.3) &  4.7 (6.8) &  7.4 (5.5) &  0.3 (1.7) &  2.2 (5.0) \\ \hline
		Cobra &  8.6 (8.6) & 25.8 (15.2) &  8.2 (8.9) &  8.5 (5.9) &  1.1 (3.5) &  6.2 (8.3) \\  
		Mixcobra &  7.7 (9.2) & 28.2 (15.4) &  5.4 (6.7) &  7.2 (5.8) &  0.3 (1.7) &  2.8 (6.2) \\ 
		\hline
	\end{tabular}
	\caption{Average classification error (and standard deviation into brackets) for the different machines and  both aggregation methods
		\textit{Cobra} and \textit{MixCobra}, computed for the simulated classification examples (1 unit $= 10^{-2}$). \label{table:classif}}
\end{table}

Concerning the aggregation procedures, \textit{Cobra} and \textit{MixCobra}  succeed in yielding very satisfactory performances.
When local information brings indirectly strong knowledge on the class label, as for the ``circles'' and ``spirals'' examples, \textit{MixCobra} outperforms
compared to \textit{Cobra}.
This may be explained by the fact that the input part of \textit{MixCobra} weights behaves like a nonparametric kernel-like method, 
performing well in such cases, as shown by the low error of the knn classifier.

\subsection{Regression}
For the regression framework, we use the eponymous R  package COBRA  available on CRAN \citep{PackCobra}
to compute the performance results for \textit{Cobra} aggregation. It should be noted that this package implements the procedure only for  the regression framework and not for classification
(so, it was not possible to use the package in the classification framework).
In the R  package COBRA, two parameters, called hereafter $\delta$ and $\gamma$, 
drive the numerical results of the  aggregation strategy (see \citealp{PackCobra}).
The smoothing parameter $\delta$ governs the consensus, insofar that it sets a threshold for the 
distance between the prediction  $r_m(x)$ for a new observation $x$  and the prediction $r_m(x_i) $
for an observation $x_i$ in the training set: for every  machine $m$,
$$ | r_m(x_i) - r_m(x) | \leq \delta. $$

The collective estimator  $T$  is  a weighted average  of the outputs of the training data set: for $x\in\R^d$,

$$ T(x)=\sum_{i=1}^n W_{i}(x) Y_i.$$ 

The value of the weight $W_i(x)$ attributed to each output observation $Y_i$, $ 1 \leq i \leq n$, of the training data set 
depends on the consensus, possibly  computed  only for a certain proportion $ \gamma$ of machines.
This approach will be referred to as ``Cobra Adaptive method'' by opposition to the ``Cobra Fixed method'' when the number of machines is fixed.

$$W_{i}(x)=\frac{\1_{\{\sum_{m=1}^p \1_{\{|r_{m}(x)-r_{m}(X_i)|\leq \delta\}}\geq p\gamma\}}}
{\sum_{j=1}^\ell {\1_{\{\sum_{m=1}^p \1_{\{|r_{m}(x)-r_{m}(X_j)|\leq \delta\}}\geq p\gamma\}}}}.$$

A too small $\delta$ value leads to many zero weights and discards a large number of data points, whereas a large value provides just an average
of all the predictions. 
The parameter $\gamma$ drives the proportion of machines to be actually considered among all initial machines.
With no a priori information on the performances of the different machines,
both parameters $\delta$ and $\gamma$ are quite hard to choose.
Practically, in the R COBRA package, these parameters are computed by cross-validation, using a subset of the observations
in the training data set. 

Concerning \textit{MixCobra}, two parameters need also to be chosen, as in the initial \textit{Cobra} method.
The $\alpha$ parameter drives the influence of the inputs (through  the proximity between a new
input observation and the input observations of the training data set) and the $\beta$ parameter drives the consensus of the predictions. It may be interesting to note that, in \textit{MixCobra}, the two parameters play more  symmetric roles than the parameters of \textit{Cobra}.
In practice, as in the COBRA  package, both parameters are chosen by cross-validation.

The following paragraphs illustrate, in two industrial applications, the benefit of 
combining several regressors relying on a consensus notion using the \textit{Cobra} and the \textit{MixCobra} method.

\paragraph{Modeling Air Compressors}
Air compressors are crucial elements of production and distribution of air gases.
Compressors are instrumented with sensors that provide valuable information on the compressor status. 
Power consumption is measured on the electric motor driving the compressor and other
information as  flow, input pressure, output pressure, cooling water temperature and air temperature
are also recorded.
Monitoring compressor power consumption is essential to an Air Separation Unit safe
and efficient operation \citep{CHM}.
In this study, $p=8$ regression machines are used to model the air compressor consumption: linear regression model (lm),
regression tree (cart), bagging regression tree (bag), random forest (rf), Support Vector Machines (svm) 
and  $k$-nearest neighbors (knn) with different values of $k$ $(k=2, 5, 10)$.

\begin{figure}[h!]
	\begin{center}
		\includegraphics[width=10cm]{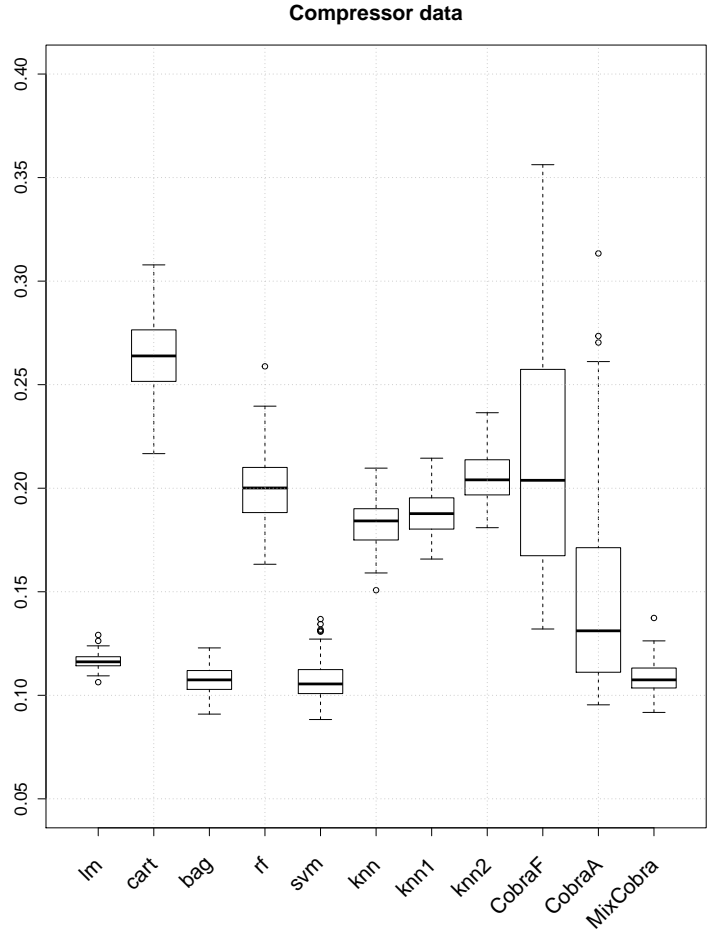}
		\caption{Performances of \textit{MixCobra} and \textit{Cobra} for a fixed or adaptive number of machines.
			The machines are lm, cart, bag, rf, svm, knn (3 machines $k=2,5,10$).}
		\label{fig:compressor}
	\end{center}  
\end{figure}

The data set contains $N=1000$ hourly observations of a working air compressor.
Each variable is first transformed to have zero mean and unit variance.
The performances are here computed by cross-validation using $K=100$ replications.
For each replication,  2/3  of the observations are randomly chosen to calibrate the machines
and 1/3 of the remaining observations are used to estimate the performances of the different machines.
The performances on the test sets are presented in Table \ref{table:compK100n1000}: for each case, the mean squared error is computed.
We observed that the cart bagging and the svm machines provide in average the smallest error.
Moreover, we observe that the best machine (i.e. the machine providing the smallest test performance error) changes from run to run.
For $K=100$ replications, the best machine 
is alternatively  the cart bagging machine (49 runs), the svm machine (49 runs)
or the lm machine (2 runs).
In this case, aggregation methods seem particularly well appropriate.
Figure \ref{fig:compressor} shows the  boxplots computed with the different machines and the 3 aggregation algorithms:
\textit{Cobra} with all machines (CobraF),
\textit{Cobra} with an adaptive number of machines (CobraA) and \textit{MixCobra}. 
As expected, the Cobra algorithm with all machines yields in average 
the worst performance among the three aggregation techniques.
Choosing adaptively the number of machines (CobraA) allows to discard a possibly bad machine and, so, improves the performance of Cobra.
We observe that \textit{MixCobra} provides, in average, the best performance, associated with a low standard deviation compared to the CobraF and CobraA aggregation methods.

\begin{table}[h!]
	\begin{center}
		\begin{tabular}{lrr}
			\hline
			& mean & std \\ 
			\hline
			lm & 11.64 & 0.00 \\ 
			cart & 26.43 & 0.02 \\ 
			bag & 10.75 & 0.01 \\ 
			rf & 20.05 & 0.02 \\ 
			svm & 10.77 & 0.01 \\ 
			knn 2& 18.35 & 0.01 \\ 
			knn 5 & 18.79 & 0.01 \\ 
			knn 10& 20.49 & 0.01 \\ \hline
			Cobra Fixed& 21.63 & 0.06 \\ 
			Cobra Adaptive & 14.81 & 0.05 \\ 
			MixCobra & 10.85 & 0.01 \\ 
			\hline
		\end{tabular}
		\caption{Average test performances of the  machines and aggregation methods for the air compressor equipment 
			(1 unit $=10^{-2}$).
			\label{table:compK100n1000}}
	\end{center}  
\end{table}

\paragraph{Modeling Wind Turbines}
The second application aims at modeling 6 different wind turbines on a wind plant in France.
Each turbine is described by 5 variables, representing half-hourly information. The production of electricity is here the target variable.
The explanatory variables are the wind power, the wind direction (sine, cosine), and the temperature. 
For each wind turbine, each variable is first transformed to have zero mean and unit variance.

Table \ref{table:windturbine} presents the performances obtained in modeling the 6 wind turbines ($W_k$, $1 \leq k \leq 6$),
for $N=1000$ operational data points, using the same $p=8$ set of learning machines as in the previous paragraph
and the corresponding \textit{Cobra} and \textit{MixCobra} aggregations.
For all wind turbines, the cart bagging machine provides in average the best performances.
However, we study in detail, for each wind turbine, the performances obtained for each run
over the $K=100$ repetitions.
Table \ref{table:windturbinerank} presents the number of runs  where every given 
machine provides the smallest error.
For the $K=100$ repetitions, the model which provides
the smallest error is never the same, and is alternatively either cart bagging or svm.
We observe that the proportion differs for each wind turbine.
In this case, aggregation methods, and particularly \textit{MixCobra}, are especially interesting,
providing a unique global approach to model all wind turbines, as repetitions show that the best model
is never the same.

\begin{table}[H]
	\begin{center}
		\begin{tabular}{lllllll}
			\hline
			& W1 & W2 & W3 & W4 & W5 & W6 \\ 
			\hline
			lm & 14.39 (0.9) & 18.22 (1.2) & 16.65 (1.2) & 15.72 (1.7) & 15.16 (1.2) & 16.97 (1.8) \\ 
			cart & 18.39 (1.7) & 19.53 (1.8) & 19.46 (1.9) & 19.10 (2.0) & 19.02 (1.7) & 19.22 (1.8) \\ 
			bag & 9.13 (0.7) & 9.95 (1.2) & 9.61 (1.0) & 9.35 (0.7) & 9.75 (0.8) & 10.17 (1.1) \\ 
			rf & 11.43 (0.9) & 16.32 (2.7) & 15.41 (2.2) & 13.89 (1.8) & 15.12 (2.0) & 14.87 (1.7) \\ 
			svm & 9.28 (1.0) & 13.55 (3.5) & 12.72 (3.5) & 11.12 (2.3) & 12.01 (2.1) & 12.00 (2.6) \\ 
			knn 2 & 13.00 (0.6) & 15.59 (1.5) & 14.65 (1.1) & 14.62 (1.0) & 15.75 (1.2) & 15.70 (1.1) \\ 
			knn 5 & 12.12 (0.7) & 15.26 (1.8) & 13.90 (1.1) & 13.60 (1.0) & 14.95 (1.3) & 14.75 (1.3) \\ 
			knn 10 & 12.69 (0.8) & 16.78 (1.8) & 15.00 (1.3) & 14.41 (1.2) & 15.93 (1.4) & 15.66 (1.5) \\ 
			\hline
			Cobra Fixed & 18.83 (4.5) & 29.55 (10.6) & 31.76 (10.9) & 24.01 (8.5) & 24.73 (9.2) & 23.92 (8.4) \\ 
			Cobra Adaptive & 14.15 (5.5) & 17.76 (8.1) & 17.38 (9.8) & 16.43 (7.4) & 17.71 (9.0) & 15.29 (7.2) \\ 
			MixCobra & 9.02 (0.6) & 10.47 (1.4) & 10.04 (1.0) & 9.45 (0.8) & 10.37 (0.9) & 10.57 (1.1) \\ 
			\hline
		\end{tabular}
		\caption{Average test  performances (and standard deviation into brackets) for the wind turbines (1 unit $=10^{-2}$).
			\label{table:windturbine}}
	\end{center}  
\end{table}

\begin{table}[H]
	\begin{center}
		\begin{tabular}{lllllll}
			\hline
			& W1 & W2 & W3 & W4 & W5 & W6 \\ 
			\hline
			lm & 0 & 0 & 0 & 0 & 0 & 0 \\ 
			cart & 0 & 0 & 0 & 0 & 0 & 0 \\ 
			bag & 52 & 89 & 88 & 76 & 93 & 77 \\ 
			rf & 0 & 0 & 0 & 0 & 0 & 0 \\ 
			svm & 48 & 11 & 12 & 24 & 7 & 23 \\ 
			knn 2 & 0 & 0 & 0 & 0 & 0 & 0 \\ 
			knn 5 & 0 & 0 & 0 & 0 & 0 & 0 \\ 
			knn 10 & 0 & 0 & 0 & 0 & 0 & 0 \\ 
			\hline
		\end{tabular}
		\caption{Number of runs  where every machine provides the best model for the wind turbines (smallest test error).
			\label{table:windturbinerank}}
	\end{center}  
\end{table}

\paragraph{Increasing the dimension or the number of estimators}

Since our combining method relies on a kernel rule, it is expected to be affected in some sense by the curse of dimensionality. However, an interesting feature of our approach is that the term based on the distance between the entries, $(X-x)/\alpha$,
and the term involving the preliminary  estimations, $(\r(X)-\r(x))/\beta$, are not affected the same way, since the ``dimension'' in not the same in both cases. Indeed, in the first case, the dimension is the actual dimension $d$ of the space $\R^d$ containing the entries,  whereas   in the second case, the role of the dimension is played by the number $p$ of individual estimators. In fact, the final combined estimator shows an interesting flexibility through the calibration of $\alpha$ and $\beta$. When the dimension $d$ increases, the method will give more weight to the combining part, which is not affected in itself by increasing $d$ and may only be affected through the particular behavior of the initial estimators considered. Conversely, for reasonable values of $d$, the effect of an increase of the number $p$ of estimators may be balanced by  the  distance term between the entries.

\begin{table}[H]
	\centering
	\begin{tabular}{rlllll}
		\hline
		& E1 $d=6$ & E2, $d=11$ & E3, $d=16$ & E4, $d=21$ & E5, $d=26$ \\ 
		\hline
		lm & 11.6 (0.4) & 11.7 (0.3) & 11.8 (0.4) & 11.9 (0.4) & 11.8 (0.3) \\ 
		cart & 26.4 (1.9) & 26.0 (2.0) & 26.6 (1.9) & 26.8 (2.2) & 26.2 (1.8) \\ 
		bag & 10.7 (0.6) & 11.4 (0.7) & 12.2 (0.7) & 12.3 (0.6) & 12.6 (0.6) \\ 
		rf & 20.1 (1.6) & 17.0 (1.3) & 17.0 (1.5) & 18.8 (1.4) & 18.2 (1.7) \\ 
		svm & 10.8 (1.0) & 14.8 (1.1) & 16.4 (0.7) & 16.6 (0.7) & 16.2 (0.7) \\ 
		knn 2 & 18.4 (1.2) & 41.3 (2.1) & 52.9 (2.3) & 60.3 (2.8) & 65.1 (3.4) \\ 
		knn 5 & 18.8 (1.0) & 36.5 (1.9) & 45.5 (1.8) & 51.9 (2.5) & 56.6 (3.1) \\ 
		knn10 & 20.5 (1.1) & 36.3 (2.1) & 44.5 (1.9) & 50.3 (2.1) & 54.4 (3.6) \\ 
		\hline
		MixCobra & 10.8 (0.7) & 13.2 (0.9) & 13.9 (0.8) & 14.5 (0.9) & 14.1 (0.4) \\ 
		\hline
	\end{tabular}
	\caption{Performances with the initial air compressor data embedded into high dimensional spaces, 
		adding respectively 5, 10, 15, 20, 25 random variables.
		\label{table:dimstudy}}
\end{table}

To study the effect  of the inputs dimensionality, the original compressor data
are embedded in successive high dimensional spaces of size $p=11, 16, 21, 26$, by artificially adding  
to the 6 initial variables $5,10,15, 20$ independent random variables uniformly distributed.

Table \ref{table:dimstudy} shows the performances computed by cross validation using $K=100$ replications.
As previously, for each replication, $2/3$ of the observations are randomly chosen
to calibrate the machines and $1/3$ of the remaining observations are used to estimate the performances
for the different machines. We observe, as expected, that the performances of the knn machines strongly deteriorate
as the dimension increases. The performances of the random forest and the svm machines are also impacted
with an increase of the dimension but to a lower extent. The performances of the lm machines and the cart machines
are stable. 
Table \ref{table:dimstudy} shows that \textit{MixCobra} performs well, even if the performances
of some machines decrease, and adapts to the increase of dimensionality.

\section{Perspectives}\label{section:ccl}

It is worth pointing out that further definitions of combining estimate linking an aggregation part and a distance between entries part could be investigated. A point of view that seem very appealing is to employ general multivariate kernels and allow the bandwidth to be different depending on the direction, that is the machine and also the direction in $\R^d$.  
The advantage of this method would be a great flexibility through his data-driven adaptive nature, whereas the increase of the number of parameters necessary for the estimation of a bandwidth matrix would represent a challenge both for estimation and computational aspects.

In our future research work, we plan to study more specifically the high dimensional case. In this framework, it could be interesting to compute more appropriate distances between entries. In particular, a promising strategy could be to try to bring into play the correlations between the output and the different explanatory variables.

\section{Proofs}\label{section:proof}

\subsection{Proof of Lemma \ref{lem:regclas}}

 For any classifier $\psi$, we may write
 \begin{align*}
\P(\psi(X)\neq Y| X=x)&=1-\P(\psi(X)= Y| X=x)\\&=1-\P(\psi(X)=1,Y=1|X=x)-
\P(\psi(X)=0, Y=0| X=x)\\&=
1-\I_{\{\psi(X)=1\}}\P(Y=1|X=x)-\I_{\{\psi(X)=0\}}\P( Y=0| X=x)\\&=
1-\I_{\{\psi(X)=1\}}\eta(x)-\I_{\{\psi(X)=0\}}(1-\eta(x)).
\end{align*}
Then, applying this to $\mathcal C_n$ and $\psi^\star$, we obtain
\begin{align*}L_n-L^\star&=\P(\C_n(X)\neq Y| X=x)-\P(\psi^\star(X)\neq Y| X=x)\\&=
1-\I_{\{\C_n(X)=1\}}\eta(x)-\I_{\{\C_n(X)=0\}}(1-\eta(x))-
1+\I_{\{\psi^\star(X)=1\}}\eta(x)+\I_{\{\psi^\star(X)=0\}}(1-\eta(x))\\&=
\eta(x)(-\I_{\{\C_n(X)=1\}}+\I_{\{\C_n(X)=0\}}+\I_{\{\psi^\star(X)=1\}}-\I_{\{\psi^\star(X)=0\}})
-\I_{\{\C_n(X)=0\}}+\I_{\{\psi^\star(X)=0\}}
\\&=2\eta(x)(\I_{\{\C_n(X)=0\}}-\I_{\{\psi^\star(X)=0\}})-\I_{\{\C_n(X)=0\}}+\I_{\{\psi^\star(X)=0\}}\\&=
(2\eta(x)-1)(\I_{\{\C_n(X)=0\}}-\I_{\{\psi^\star(X)=0\}}).
\end{align*}
This implies
\begin{align*}
L_n-L^\star=\int|2\eta(x )-1|\I_{\{\psi^\star(x)\neq \C_n(x)\}}\mu(dx).
\end{align*}
The event $\{\psi^\star(x)\neq \C_n(x)\}$ corresponds to either one of the two following situations:
\begin{itemize}
\item $\psi^\star(x)=0$, $\C_n(x)=1$, which can be rewritten $\eta(x)\leq 1/2$ and  $2\T^\star_n(x)-\frac{\sum_{i=1}^n  K_\alpha(Z_i-z)}{n\E [ K_\alpha(Z_i-z)]}>0$.
In this case, we have \begin{align*}|2\eta(x)-1|&=1-2\eta(x)\\&=1- \frac{\sum_{i=1}^n K_\alpha(Z_i-z)}{n\E  [K_\alpha(Z-z)]}+ \frac{\sum_{i=1}^n K_\alpha(Z_i-z)}{n\E [ K_\alpha(Z-z)]}-2\eta(x)\\&< 1- \frac{\sum_{i=1}^n K_\alpha(Z_i-z)}{n\E  [K_\alpha(Z-z)]}+2\T^\star_n(x)-2\eta(x)\\&\leq \left|1-\eta(x)-\frac{\sum_{i=1}^n K_\alpha(Z_i-z)}{n\E  [K_\alpha(Z-z)]}+\T^\star_n(x)\right|+|\eta(x)-\T^\star_n(x)|.\end{align*}
\item $\psi^\star(x)=1, \C_n(x)=0$, which means that $\eta(x)> 1/2$ and $2\T_n^\star(x)-\frac{\sum_{i=1}^n  K_\alpha(Z_i-z)}{n\E  [K_\alpha(Z-z)]}\leq 0$.
We have \begin{align*}|2\eta(x)-1|&=2\eta(x)-1\\&=2\eta(x)- \frac{\sum_{i=1}^n K_\alpha(Z_i-z)}{n\E [ K_\alpha(Z-z)]}+ \frac{\sum_{i=1}^n K_\alpha(Z_i-z)}{n\E [ K_\alpha(Z-z)]}-1\\&< 2\eta(x)+ \frac{\sum_{i=1}^n K_\alpha(Z_i-z)}{n\E  [K_\alpha(Z-z)]}-2\T^\star_n(x)-1\\&\leq \left|1-\eta(x)-\frac{\sum_{i=1}^n K_\alpha(Z_i-z)}{n\E  [K_\alpha(Z-z)]}+\T^\star_n(x)\right|+|\eta(x)-\T^\star_n(x)|.\end{align*}

\end{itemize}

Hence, $$L_n-L^\star\leq \int\left|1-\eta(x)-\frac{\sum_{i=1}^n  K_\alpha(Z_i-z)}{n\E [ K_\alpha(Z-z)]}+\T_n^\star(x)\right|\mu(dx)+\int|\eta(x)-\T_n^\star(x)|\mu(dx).$$

%
%
%

\subsection{Proof of the covering lemma \ref{lem:cov}}

\begin{enumerate}
\item 
We recover $\R^{d+p}$ with balls with radius $\rho/2$ and centers $v_1,\dots, v_j,\dots$.
Each element of $\R^{d+p}$ is in a finite number of such balls. This number depends on the dimension $d+p$.

For the first bound, we search for an upper bound for $g\left(\frac{u-x}{\alpha},
\frac{\f(u)-\f(x)}{\beta}\right)$, and a lower bound for $\E\left[g\left(\frac{X-x}{\alpha},
\frac{\f(X)-\f(x)}{\beta}\right)\right]$ for every $x$ such that 
$\left\|\frac{z-t}{\alpha}-v_j\right\|\leq \frac{\rho}{2}$. Here, we use the notation $z=\left(x_{1},\dots, x_{d},\frac\alpha\beta f_1(x)\dots,\frac\alpha\beta f_p(x)\right)$ and $t=\left(u_{1},\dots, u_{d},\frac\alpha\beta f_1(u)\dots,\frac\alpha\beta f_p(u)\right)$.

For the upper bound, we note that the element of $\R^{d+p}$ $$\frac{z-t}{\alpha}=\left(\frac{x-u}\alpha,\frac{\f(x)-\f(u)}\beta\right)=\frac 1\alpha\left(x_1-u_1,\dots,x_d-u_d,\frac \alpha\beta f_1(x)-\frac \alpha\beta f_1(u),\dots,\frac \alpha\beta f_p(x)-\frac \alpha\beta f_p(u)\right) $$ belongs to a finite number of balls of radius $\rho/2$ and center one of the $v_j$'s. We are going to upper bound the quantity of interest by the sum of the suprema over these different balls.
The fact that $\frac{z-t}{\alpha}$ belongs to such a ball  (with a center $v_j$) means that $\left\|\frac{z-t}{\alpha}-v_j\right\|\leq \frac \rho 2$.
We may write \begin{align*}
&g\left(\frac{u-x}{\alpha},
\frac{\f(u)-\f(x)}{\beta}\right)\\&\qquad\leq
\sum_{j=1}^{+\infty}\sup_{x',\left\|\frac{z'-t}{\alpha}-v_j\right\|\leq \frac \rho 2}g\left(\frac{u-x'}{\alpha},
\frac{\f(u)-\f(x')}{\beta}\right)
\I_{\left\{\left\|\frac{z-t}{\alpha}-v_j\right\|\leq \frac \rho 2\right\}}(x).
\end{align*}
For the lower bound, we use the property $K(z)\geq c\I_{B(0,\rho)}(z)$. We have
\begin{align*}\E \left[g\left(\frac{X-x}{\alpha},
\frac{\f(X)-\f(x)}{\beta}\right)\right]
&=\E \left[K\left(\frac{z-Z}\alpha\right)\right]\\&\geq c\E\left[\I_{\left\{\left\|\frac{z-Z}\alpha\right\|\leq \rho\right\}}\right]\\&\geq
c\P\bigg(\left\|\frac{z-Z}\alpha\right\|\leq \rho\bigg).
\end{align*}
Now, if $\left\|\frac{z-t}{\alpha}-v_j\right\|\leq \frac{\rho}{2}$, let us check that $$\left\|\frac {Z-t}{\alpha}-v_j\right\|\leq \frac{\rho}{2} $$
implies that 

$$\left\|\frac {Z-z}{\alpha}\right\|\leq \rho.$$
We have
\begin{align*}
\left\|\frac{z-Z}{\alpha}\right\|&=\left\|\frac{z-t+t-Z}{\alpha}-v_j+v_j\right\|
\\&\leq \left\|\frac{z-t}{\alpha}-v_j\right\|+\left\|\frac{t-Z}{\alpha}+v_j\right\|\\&\leq \rho.
\end{align*}
Thus, when $\left\|\frac{z-t}{\alpha}-v_j\right\|\leq \frac \rho 2$, 
 we obtain
$$\E  \left[g\left(\frac{X-x}{\alpha},
\frac{\f(X)-\f(x)}{\beta}\right)\right]\geq c\P\left(\left\|\frac{Z-t}{\alpha}-v_j\right\|\leq \frac{\rho}{2}\right).$$

Consequently, we have
\begin{align*}&\int  \frac{g\left(\frac{u-x}{\alpha},
\frac{\f(u)-\f(x)}{\beta}\right)}
{\E \left[g\left(\frac{X-x}{\alpha},
\frac{\f(X)-\f(x)}{\beta}\right)\right]}\mu(dx)\\&\leq \frac 1 c\sum_{j=1}^{+\infty}\frac{\int\sup_{x',\left\|\frac{z'-t}{\alpha}-v_j\right\|\leq \frac{\rho}{2}}g\left(\frac{u-x'}{\alpha},
\frac{\f(u)-\f(x')}{\beta}\right)
\I_{\left\{\left\|\frac{z-t}{\alpha}-v_j\right\|\leq \frac{\rho}{2}\right\}}(x)\mu(dx)}{\P\left(\left\|\frac{Z-t}{\alpha}-v_j\right\|\leq \frac{\rho}{2}\right)}
\\&\leq \frac 1 c\sum_{j=1}^{+\infty}\frac{\sup_{x',\left\|\frac{z'-t}{\alpha}-v_j\right\|\leq \frac{\rho}{2}}g\left(\frac{u-x'}{\alpha},
\frac{\f(u)-\f(x')}{\beta}\right)
\P\left(\left\|\frac{Z-t}{\alpha}-v_j\right\|\leq \frac{\rho}{2}\right)}{\P\left(\left\|\frac{Z-t}{\alpha}-v_j\right\|\leq \frac{\rho}{2}\right)}
\\&\leq \frac 1 c\sum_{j=1}^{+\infty}\sup_{x',\left\|\frac{z'-t}{\alpha}-v_j\right\|\leq \frac{\rho}{2}}g\left(\frac{u-x'}{\alpha},
\frac{\f(u)-\f(x')}{\beta}\right)
\\&= \frac 1 c\sum_{j=1}^{+\infty}\sup_{x',\left\|\frac{z'-t}{\alpha}-v_j\right\|\leq \frac{\rho}{2}}K\left(\frac{t-z'}{\alpha}\right)
\\&\leq \frac 1 c\sum_{j=1}^{+\infty}\sup_{s\in B(v_j,\rho/2)}K(s)
\\&\leq \frac 1 c\frac{1}{\lambda_{d+p}\big(B(0,\rho/2)\big)}\sum_{j=1}^{+\infty}\int_{B(v_j,\rho/2)}\sup_{s\in B(v_j,\rho/2)}K(s)dt
\\&\leq \frac 1 c\frac{1}{\lambda_{d+p}\big(B(0,\rho/2)\big)}\sum_{j=1}^{+\infty}\int_{ B(v_j,\rho/2)}\sup_{s\in B(t,\rho)}K(s)dt
\\&\leq \frac 1 c\frac{k_{d+p}}{\lambda_{d+p}\big(B(0,\rho/2)\big)}\int \sup_{s\in B(t,\rho)}K(s)dt = R.
\end{align*}
Here, $\lambda_{d+p}$ denotes the Lebesgue measure of $\R^{d+p}$ and $k_{d+p}$ stands for the number of covering balls containing a certain element of $\R^{d+p}$: integrating over all such balls consists in integrating over  $\R^{d+p}$ and multiplying by the number of balls $k_{d+p}$.

Finally, we use that the kernel is regular.

\bigskip

\item
For the second result, we use the same upper and lower bounds.
 
We  write 
\begin{align*}
&g\left(\frac{u-x}{\alpha},
\frac{\f(u)-\f(x)}{\beta}\right)\I_{\{\|x-u\|\geq \delta\}}\\&\leq
\sum_{j=1}^{+\infty}\sup_{x',\left\|\frac{z'-t}{\alpha}-v_j\right\|
\leq \frac{\rho}{2}}g\left(\frac{u-x'}{\alpha},
\frac{\f(u)-\f(x')}{\beta}\right)\I_{\{\|x'-u\|\geq \delta\}}
\I_{\left\{\left\|\frac{z-t}{\alpha}-v_j\right\|\leq \frac{\rho}{2}\right\}}(x).
\end{align*}
As above, when $\left\|\frac{z-t}{\alpha}-v_j\right\|\leq \frac \rho 2$,
$$\E \left[g\left(\frac{X-x}{\alpha},
\frac{\f(X)-\f(x)}{\beta}\right)\right]\geq c\P\left(\left\|\frac{Z-t}{\alpha}-v_j\right\|\leq \frac{\rho}{2}\right).$$
Then, 
\begin{align*}&\int  \frac{g\left(\frac{u-x}{\alpha},
\frac{\f(u)-\f(x)}{\beta}\right)\I_{\{\|x-u\|\geq \delta\}}}
{\E \left[g\left(\frac{X-x}{\alpha},
\frac{\f(X)-\f(x)}{\beta}\right)\right]}\mu(dx)\\&\leq \frac 1 c\sum_{j=1}^{+\infty}\frac{\int \sup_{x',\left\|\frac{z'-t}{\alpha}-v_j\right\|
\leq \frac{\rho}{2}}g\left(\frac{u-x'}{\alpha},
\frac{\f(u)-\f(x')}{\beta}\right)\I_{\{\|x'-u\|\geq \delta\}}
\I_{\left\{\left\|\frac{z-t}{\alpha}-v_j\right\|\leq \frac{\rho}{2}\right\}}(x)\mu(dx)}{\P\left(\left\|\frac{Z-t}{\alpha}-v_j\right\|\leq \frac{\rho}{2}\right)}
\\&\leq \frac 1 c\sum_{j=1}^{+\infty}\frac{\sup_{x',\left\|\frac{z'-t}{\alpha}-v_j\right\|
\leq \frac{\rho}{2}}g\left(\frac{u-x'}{\alpha},
\frac{\f(u)-\f(x')}{\beta}\right)\I_{\{\|x'-u\|\geq \delta\}}\P\left(\left\|\frac{Z-t}{\alpha}-v_j\right\|\leq \frac{\rho}{2}\right)
}{\P\left(\left\|\frac{Z-t}{\alpha}-v_j\right\|\leq \frac{\rho}{2}\right)}
\\&\leq \frac 1 c\sum_{j=1}^{+\infty}\sup_{x',\left\|\frac{z'-t}{\alpha}-v_j\right\|
\leq \frac{\rho}{2}}g\left(\frac{u-x'}{\alpha},
\frac{\f(u)-\f(x')}{\beta}\right)\I_{\{\|x'-u\|\geq \delta\}}
\\&\leq \frac 1 c\sum_{j=1}^{+\infty}\sup_{s\in B(v_j,\rho/2)}K(s)\I_{\{\|s_1\|\geq \delta/\alpha\}},
\end{align*}
where $s_1\in\R^d$ denotes the vector of the $d$ first coordinates of $s\in\R^{d+p}$. 
Let us check that this quantity converges to 0 as $\alpha$ tends to 0.
Indeed, it is a series with nonnegative terms, of the type $$\sum_{j=1}^{+\infty}\varphi_j(\alpha),$$ which is finite  for every $\alpha$ thanks to the first part of the covering lemma. Since  $\lim_{\alpha\to 0} \varphi_j(\alpha)=0$ for every $j\geq 1$, the series converges to 0.
\end{enumerate}

\subsection{Proof of Proposition \ref{prop:dec}}

\begin{enumerate}
\item Since $f$ is supposed to be continuous with compact support, it is bounded and uniformly continuous on its support: there exists $\delta >0$ such that  $\|x-u\|<\delta$ implies $|f(x)-f(u)|<\e '$.
We have \begin{align*}
& \int\left|f(x)-\frac{\E\left[ f(X)g\left(\frac{X-x}{\alpha},
\frac{\f(X)-\f(x)}{\beta}\right)\right]}{\E \left[g\left(\frac{X-x}{\alpha},
\frac{\f(X)-\f(x)}{\beta}\right)\right]}\right|\mu(dx)
\\&=\int\left|
f(x)-\frac{\int f(u)g\left(\frac{u-x}{\alpha},
\frac{\f(u)-\f(x)}{\beta}\right)\mu(du)}{\E \left[g\left(\frac{X-x}{\alpha},
\frac{\f(X)-\f(x)}{\beta}\right)\right]
}
\right|
\mu(dx)
\\&=\int\left|\frac{
f(x)\E \left[g\left(\frac{X-x}{\alpha},
\frac{\f(X)-\f(x)}{\beta}\right)\right]-\int f(u)g\left(\frac{u-x}{\alpha},
\frac{\f(u)-\f(x)}{\beta}\right)\mu(du)}{\E \left[g\left(\frac{X-x}{\alpha},
\frac{\f(X)-\f(x)}{\beta}\right)\right]
}
\right|
\mu(dx)
\\&=\int\left|\frac{
f(x)\int g\left(\frac{u-x}{\alpha},
\frac{\f(u)-\f(x)}{\beta}\right)\mu(du)-\int f(u)g\left(\frac{u-x}{\alpha},
\frac{\f(u)-\f(x)}{\beta}\right)\mu(du)}{\E \left[g\left(\frac{X-x}{\alpha},
\frac{\f(X)-\f(x)}{\beta}\right)\right]
}
\right|
\mu(dx)
\\&=\int\left|\frac{
\int g\left(\frac{u-x}{\alpha},
\frac{\f(u)-\f(x)}{\beta}\right)(f(x)-f(u))\mu(du)}{\E \left[g\left(\frac{X-x}{\alpha},
\frac{\f(X)-\f(x)}{\beta}\right)\right]
}
\right|
\mu(dx)\\
\\&\leq\int\int\frac{
 g\left(\frac{u-x}{\alpha},
\frac{\f(u)-\f(x)}{\beta}\right)}{\E \left[g\left(\frac{X-x}{\alpha},
\frac{\f(X)-\f(x)}{\beta}\right)\right]
}|f(x)-f(u)|\mu(du)
\mu(dx)\\&=
\int \int_{B(x,\delta)}\frac{
 g\left(\frac{u-x}{\alpha},
\frac{\f(u)-\f(x)}{\beta}\right)}{\E \left[g\left(\frac{X-x}{\alpha},
\frac{\f(X)-\f(x)}{\beta}\right)\right]
}|f(x)-f(u)|\mu(du)\mu(dx)\\&+
\int \int\frac{
 g\left(\frac{u-x}{\alpha},
\frac{\f(u)-\f(x)}{\beta}\right)}{\E \left[g\left(\frac{X-x}{\alpha},
\frac{\f(X)-\f(x)}{\beta}\right)\right]
}\I_{\{\|x-u\|\geq \delta\}}(u)|f(x)-f(u)|\mu(du)\mu(dx)
\end{align*}
The first term may be upper bounded by $\e '$ by uniform continuity of $f$, since, for every $x$,  $$\int_{B(x,\delta)}\frac{
 g\left(\frac{u-x}{\alpha},
\frac{\f(u)-\f(x)}{\beta}\right)}{\E \left[g\left(\frac{X-x}{\alpha},
\frac{\f(X)-\f(x)}{\beta}\right)\right]
}\mu(dy)\leq 1.$$
For the second term, we first use  $|f(x)-f(u)|\leq 1$, then, the second point in the covering lemma \ref{lem:cov} implies  that the integral against $\mu(du)$ tends to 0, and the convergence to 0 of the integral against $\mu(dx)$ follows  from Lebesgue's dominated convergence theorem, since the integrand is bounded by $R$ according to the first point in the covering lemma \ref{lem:cov}. Thus, we get
$$\limsup_{n\to\infty}\int\left|f(x)-\frac{\E\left[ f(X)g\left(\frac{X-x}{\alpha},
\frac{\f(X)-\f(x)}{\beta}\right)\right]}{\E \left[g\left(\frac{X-x}{\alpha},
\frac{\f(X)-\f(x)}{\beta}\right)\right]}\right|\mu(dx)\leq \e ' .$$

\item We have \begin{align*}&\int\left|\frac{\E\left[ f(X)g\left(\frac{X-x}{\alpha},
\frac{\f(X)-\f(x)}{\beta}\right)\right]}{\E \left[g\left(\frac{X-x}{\alpha},
\frac{\f(X)-\f(x)}{\beta}\right)\right]}-\E [\T^\star_n(x)]\right|\mu(dx)\\&=
\int\left|\frac{\int f(u)g\left(\frac{u-x}{\alpha},
\frac{\f(u)-\f(x)}{\beta}\right)\mu(du)}{\E \left[g\left(\frac{X-x}{\alpha},
\frac{\f(X)-\f(x)}{\beta}\right)\right]}-\E [\T^\star_n(x)]\right|\mu(dx)
\\&=
\int\left|\frac{\int f(u)g\left(\frac{u-x}{\alpha},
\frac{\f(u)-\f(x)}{\beta}\right)\mu(du)}{\E \left[g\left(\frac{X-x}{\alpha},
\frac{\f(X)-\f(x)}{\beta}\right)\right]}-\frac{\E\left[\sum_{i=1}^n Y_i g\left(\frac{X_i-x}{\alpha},
\frac{\f(X_i)-\f(x)}{\beta}\right)\right]}{n\E\left[g\left(\frac{X-x}{\alpha},
\frac{\f(X)-\f(x)}{\beta}\right)\right]}\right|\mu(dx)
\\&=
\int\left|\frac{\int f(u)g\left(\frac{u-x}{\alpha},
\frac{\f(u)-\f(x)}{\beta}\right)\mu(du)-\E\left[\eta( X) g\left(\frac{X-x}{\alpha},
\frac{\f(X)-\f(x)}{\beta}\right)\right]}{\E\left[g\left(\frac{X-x}{\alpha},
\frac{\f(X)-\f(x)}{\beta}\right)\right]}\right|\mu(dx)
\\&=
\int\left|\frac{\int (f(u)-\eta(u))g\left(\frac{u-x}{\alpha},
\frac{\f(u)-\f(x)}{\beta}\right)\mu(du)}{\E\left[g\left(\frac{X-x}{\alpha},
\frac{\f(X)-\f(x)}{\beta}\right)\right]}\right|\mu(dx)\\&\leq
\int\int |f(u)-\eta(u)| \frac{g\left(\frac{u-x}{\alpha},
\frac{\f(u)-\f(x)}{\beta}\right)}{\E\left[g\left(\frac{X-x}{\alpha},
\frac{\f(X)-\f(x)}{\beta}\right)\right]}\mu(du)\mu(dx)\\&=
\int\int  \frac{g\left(\frac{u-x}{\alpha},
\frac{\f(u)-\f(x)}{\beta}\right)}{\E\left[g\left(\frac{X-x}{\alpha},
\frac{\f(X)-\f(x)}{\beta}\right)\right]}\mu(dx)|f(u)-\eta(u)|
\mu(du)\\
&\leq \int R |f(u)-\eta(u)|\mu(du)\leq R \e '. 
\end{align*} Here, we used the Fubini Theorem, then the covering lemma, and finally the definition of  $f$.
\item For the last term $\int\E\left[|\E [\T^\star_n(x)]-\T^\star_n(x)|\right]\mu(dx)$, 
let us write, for every $x$, 
\begin{align*}
\left(\E\left[|\E [\T^\star_n(x)]-\T^\star_n(x)|\right]\right)^2&\leq \E\left[|\E [\T^\star_n(x)]-\T^\star_n(x)|^2\right]
\\&=\E\left[\left|\frac{\frac 1 n\sum_{i=1}^n \E \left[Y_i g\left(\frac{X_i-x}{\alpha},
\frac{\f(X_i)-\f(x)}{\beta}\right)\right]}{\E\left[g\left(\frac{X-x}{\alpha},
\frac{\f(X)-\f(x)}{\beta}\right)\right]}-  \frac{\frac 1 n\sum_{i=1}^n Y_i g\left(\frac{X_i-x}{\alpha},
\frac{\f(X_i)-\f(x)}{\beta}\right)}{\E\left[g\left(\frac{X-x}{\alpha},
\frac{\f(X)-\f(x)}{\beta}\right)\right]} \right|^2\right]
\\&=
\frac{\E\left[\left|
\frac 1 n\sum_{i=1}^n
\left(
\E \left[Y_i g\left(\frac{X_i-x}{\alpha},
\frac{\f(X_i)-\f(x)}{\beta}\right)\right]
- Y_i g\left(\frac{X_i-x}{\alpha},
\frac{\f(X_i)-\f(x)}{\beta}\right)\right)\right|^2\right]
}{\left(\E\left[g\left(\frac{X-x}{\alpha},
\frac{\f(X)-\f(x)}{\beta}\right)\right]\right)^2
}
\\&=
\frac{\E\left[\left|
\sum_{i=1}^n
\left(
\E \left[Y g\left(\frac{X-x}{\alpha},
\frac{\f(X)-\f(x)}{\beta}\right)\right]
- Y_ig\left(\frac{X_i-x}{\alpha},
\frac{\f(X_i)-\f(x)}{\beta}\right)\right)\right|^2\right]
}{n^2\left(\E\left[g\left(\frac{X-x}{\alpha},
\frac{\f(X)-\f(x)}{\beta}\right)\right]\right)^2
}
\end{align*}
Using that the variance of a sum of $n$  i.i.d variables is $n$ times the variance of one such variable, and then upper bounding this variance by the expectation of the squared variable, we obtain, for every $x$,
\begin{align*}\label{eq:var}
&\left(\E[|\E [\T^\star_n(x)]-\T^\star_n(x)|]\right)^2\\&\leq
\frac{n\E\left[
\left(
\E \left[Y g\left(\frac{X-x}{\alpha},
\frac{\f(X)-\f(x)}{\beta}\right)\right]
- Y g\left(\frac{X-x}{\alpha},
\frac{\f(X)-\f(x)}{\beta}\right)\right)^2\right]
}{n^2\left(\E\left[g\left(\frac{X-x}{\alpha},
\frac{\f(X)-\f(x)}{\beta}\right)\right]\right)^2
}\\&=
\frac{\E\left[
\left(
\E \left[Y g\left(\frac{X-x}{\alpha},
\frac{\f(X)-\f(x)}{\beta}\right)\right]
- Y g\left(\frac{X-x}{\alpha},
\frac{\f(X)-\f(x)}{\beta}\right)\right)^2\right]
}{n\left(\E\left[g\left(\frac{X-x}{\alpha},
\frac{\f(X)-\f(x)}{\beta}\right)\right]\right)^2
}
\\&\leq
\frac{\E\left[\left( Y g\left(\frac{X-x}{\alpha},
\frac{\f(X)-\f(x)}{\beta}\right)\right)^2\right]
}{n\left(\E\left[g\left(\frac{X-x}{\alpha},
\frac{\f(X)-\f(x)}{\beta}\right)\right]\right)^2
}
\\&\leq
\frac{\E\left[ g\left(\frac{X-x}{\alpha},
\frac{\f(X)-\f(x)}{\beta}\right)^2\right]
}{n\left(\E\left[g\left(\frac{X-x}{\alpha},
\frac{\f(X)-\f(x)}{\beta}\right)\right]\right)^2
},
\end{align*} since $Y$ is bounded by 1.
Let $G$ denote an upper bound on $g$, which exists because the kernel $K$ is regular. Thus, for every $x$, 
$ g\left(\frac{X-x}{\alpha},
\frac{\f(X)-\f(x)}{\beta}\right)\leq G$, so that $$\frac 1 G g\left(\frac{X-x}{\alpha},
\frac{\f(X)-\f(x)}{\beta}\right)\leq 1.$$ Consequently, we get $$\frac{g\left(\frac{X-x}{\alpha},
\frac{\f(X)-\f(x)}{\beta}\right)^2}{G^2}\leq\frac{g\left(\frac{X-x}{\alpha},
\frac{\f(X)-\f(x)}{\beta}\right)}{G} .$$ Then $$\frac{\E \left[g\left(\frac{X-x}{\alpha},
\frac{\f(X)-\f(x)}{\beta}\right)^2\right]}{G^2}\leq\frac{\E  \left[g\left(\frac{X-x}{\alpha},
\frac{\f(X)-\f(x)}{\beta}\right)\right]}{G}, $$ and thus,
$$\frac{\E \left[g\left(\frac{X-x}{\alpha},
\frac{\f(X)-\f(x)}{\beta}\right)^2\right]}{\E  \left[g\left(\frac{X-x}{\alpha},
\frac{\f(X)-\f(x)}{\beta}\right)\right]}\leq{G} .$$
Applying this result to the inequality above  yields 
\begin{align*}
\left(\E[|\E [\T^\star_n(x)]-\T^\star_n(x)|]\right)^2&\leq
\frac{G
}{n\E  \left[g\left(\frac{X-x}{\alpha},
\frac{\f(X)-\f(x)}{\beta}\right)\right]
}
\\&\leq
\frac{G
}{n c\P(\|\frac{z-Z}{\alpha}\|\leq \rho)
}
,
\end{align*} with  $Z=(X_{1},\dots, X_{d},\frac\alpha\beta f_1(X),\dots,\frac\alpha\beta f_p(X))$ and $z=(x_{1},\dots, x_{d},\frac\alpha\beta f_1(x)\dots,\frac\alpha\beta f_p(x))$.
We are interested in the integral  against $\mu(dx)$ of this quantity.

Let us divide this integral between an integral over a cube  of side $2\ell$ centered at the origin $\mathcal Q=[-\ell,\ell]^d$ and an integral over the complement of this  cube. We may write
$$\int_{\mathcal Q^c} \E[|\E [\T^\star_n(x)]-\T^\star_n(x)|]\mu(dx) \leq 2\int_{\mathcal Q^c}\E [\T^\star_n(x)]\mu(dx).$$ This quantity converges to $$2\int_{\mathcal Q^c}\eta(x)\mu(dx).$$ Indeed, the upper bounds obtained for the integral against $\mu(dx)$ of the first 3 terms in \eqref{eq:decoup} remain true when integrating over  $\mathcal Q^c\subset \R^d$, and they ensure the convergence of $\int_{\mathcal Q^c}\E [\T^\star_n(x)]\mu(dx)$
to $\int_{\mathcal Q^c}\eta(x)\mu(dx)$. Moreover, the quantity $2\int_{\mathcal Q^c}\eta(x)\mu(dx)$ can be made arbitrarily small by choosing an appropriate value of $\ell$, because $\left|2\int_{\mathcal Q^c}\eta(x)\mu(dx)\right|\leq 2\mu(\mathcal Q^c)$ since $|Y|\leq 1.$ So, for $\ell$ and $n$  large enough, we have $$\int_{\mathcal Q^c} \E[|\E [\T^\star_n(x)]-\T^\star_n(x)|]\mu(dx)\leq\e '.$$

For the integral over $\mathcal Q=[-\ell,\ell]^d$, we will recover by balls $B(v_j,\frac{r\alpha}2)$ a parallelepiped included in $\R^{d+p}$, containing all the elements of the form $( x_1,\dots, x_d,\frac \alpha\beta f_1(x),\dots,\frac \alpha\beta f_p(x))$, with  $x\in [-\ell,\ell]^d:$
$$\left\{s\in\R^{d+p}, (s_1,\dots,s_d)\in[-\ell,\ell]^d  ,(s_{d+1},\dots,s_{d+p})\in \left[-\frac\alpha\beta,\frac\alpha\beta\right]^p\right\}.$$ This set is the parallelepiped centered at the origin with $d$ sides of length $2\ell $ and $p$ sides of length $2\frac\alpha\beta$. Let $\lceil x\rceil$ denote the ceil of $x$. For  a covering, the following number of balls is needed:
 $$N=\left\lceil\frac{4\ell}{r\alpha }\right\rceil^d\left\lceil\frac{4}{r\beta}\right\rceil^p\leq\left(\frac{8\ell}{r\alpha }\right)^d\left(\frac{8}{r\beta}\right)^p=\frac{8^{d+p}\ell^d}{r^{d+p}\alpha^d\beta^p}.$$
We have \begin{align*}
&\int\frac 1{\sqrt{\P(\|\frac{z-Z}{\alpha}\|\leq \rho)}}\mu(dx)\\&\leq
\sum_{j=1}^N\int \frac{\I_{\left\{\left\|\frac{z-v_j}{\alpha}\right\|\leq \frac{ \rho}2\right\}}(x)}{\sqrt{\P(\|\frac{z-Z}{\alpha}\|\leq \rho)}}\mu(dx)
\end{align*}
Let us check that, if $\left\|\frac{z-v_j}{\alpha}\right\|\leq \frac \rho2$, then  $\left\|\frac{Z-v_j}{\alpha}\right\|\leq \frac \rho2 $ implies $ \left\|\frac{z-Z}{\alpha}\right\|\leq  \rho$.
Indeed, 
\begin{align*}
\left\|\frac {z-Z}\alpha \right\|&\leq \left\|\frac {z-v_j}\alpha \right\|+\left\|\frac {v_j-Z}\alpha \right\|\\&\leq \frac \rho 2+\frac \rho 2=\rho.
\end{align*}

Thus, \begin{align*}
&\int\frac 1{\sqrt{\P\left(\left\|\frac{z-Z}{\alpha}\right\|\leq \rho\right)}}\mu(dx)\\&\leq
\sum_{j=1}^N\int \frac{\I_{\left\{\left\|\frac{z-v_j}{\alpha}\right\|\leq \frac{ \rho}2\right\}}(x)}{\sqrt{\P\left(\left\|\frac{Z-v_j}{\alpha}\right\|\leq \frac{ \rho}2\right)}}\mu(dx)
\\&\leq
\sum_{j=1}^N {\sqrt{\P\left(\left\|\frac{Z-v_j}{\alpha}\right\|\leq \frac{ \rho}2\right)}}
\\&\leq
\sqrt{N\sum_{j=1}^N {\P\left(\left\|\frac{Z-v_j}{\alpha}\right\|\leq \frac{ \rho}2\right)}}\quad \mbox{(Cauchy-Schwarz)}
\\&\leq \sqrt{\frac{8^{d+p}\ell^{d}}{r^{{d+p}}\alpha^{d}\beta^{p}}\times k_{p+d}},
\end{align*}
where $k_{p+d}$ is the number of balls containing an element of $\R^{d+p}$. Indeed, the sum is over all balls, so that an upper bound is obtained by multiplying the maximal measure, equal to 1, by the number of balls containing one element.

Then, we obtain
\begin{align*}\int_{\mathcal Q} \E[|\E [\T^\star_n(x)]-\T^\star_n(x)|]\mu(dx) &\leq \sqrt{\frac{G}{nc}}\sqrt{\frac{8^{d+p}\ell^{d}}{r^{{d+p}}\alpha^{d}\beta^{p}}\times k_{p+d}}\\&\leq C(n\alpha^d\beta^p)^{-1/2},\end{align*} where $C\geq 0$ is a constant depending on $\ell$, the kernel $K$ and the dimensions $d$ and $p$. 
Finally, we have proved that for $n$ sufficiently large, $$\int \E[|\E [\T^\star_n(x)]-\T^\star_n(x)|]\mu(dx) \leq \e '+ C(n\alpha^d\beta^p)^{-1/2}.$$
\end{enumerate}

\subsection{Proof of Lemma \ref{lem:etaTn*}}
According to equation \eqref{eq:dec2}, we have, for every $\e>0$, 
\begin{multline}\P\left(\int |\eta(x)-\T^\star_n(x)|\mu(dx)\geq \e/2\right)\\\leq \P\left(\int \E[|\eta(x)-\T^\star_n(x)|]\mu(dx)\geq \e/4\right)\\+\P\left(\int |\eta(x)-\T^\star_n(x)|\mu(dx)-\int \E[|\eta(x)-\T^\star_n(x)|]\mu(dx)\geq \e/4\right).\end{multline}
Thanks to Equation \eqref{eq:decoup} and Proposition \ref{prop:dec}, provided that $\alpha$ tends to 0 and $n\alpha^d\beta^p$ tends to $\infty$, the quantity $\int\E[|\eta(x)-\T^\star_n(x)|]\mu(dx)$ tends to 0 as $n$ tends to $\infty$.
So, for $n$ large enough, \begin{align*}\P\left(\int|\eta(x)-\T^\star_n(x)|\mu(dx)\geq \e/2\right)&\leq \P\left(\int|\eta(x)-\T^\star_n(x)|\mu(dx)-\int \E[|\eta(x)-\T^\star_n(x)|]\mu(dx)\geq \e/4\right)
\\&\leq \P\left(\int|\eta(x)-\T^\star_n(x)|\mu(dx)-\E \left[\int|\eta(x)-\T^\star_n(x)|\mu(dx)\right]\geq \e/4\right).\end{align*}
For controlling this quantity, 
 McDiarmid's inequality will be used.
Let $t_n^\star(x)$ denote $\T^\star_n(x)$ for values $(x_1,\dots,x_n)$ and ${t^\star_n}'(x)$ be the version of $t^\star_n(x)$ where $(x_i,y_i)$ has been replaced by $(x'_i,y'_i)$. We compute 
\begin{align*}
\int|\eta(x)-t^\star_n(x)|\mu(dx)- \int|\eta(x)-{t^\star_n}'(x)|\mu(dx)
&\leq \int|t_n^\star(x)-{t^\star_n}'(x)|\mu(dx)\\&\leq
\frac{ \int\left| y_i g\left(\frac{x_i-x}{\alpha},
\frac{\f(x_i)-\f(x)}{\beta}\right)-y_i' g\left(\frac{x_i'-x}{\alpha},
\frac{\f(x_i')-\f(x)}{\beta}\right)\right|\mu(dx)}{n\E \left[g\left(\frac{X-x}{\alpha},
\frac{\f(X)-\f(x)}{\beta}\right)\right]}
\\&\leq
2\frac{ \sup_u\int g\left(\frac{u-x}{\alpha},
\frac{\f(u)-\f(x)}{\beta}\right)\mu(dx)}{n\E \left[g\left(\frac{X-x}{\alpha},
\frac{\f(X)-\f(x)}{\beta}\right)\right]}\\&\leq \frac{2R}n.
\end{align*} 
Then, McDiarmid's inequality implies
$$\P\left(\int|\eta(x)-\T^\star_n(x)|\mu(dx)-\E \left[\int|\eta(x)-\T^\star_n(x)|\mu(dx)\right]\geq \e/4\right)\leq \exp\left(-\frac{2\e^2}{16\sum_{i=1}^n(\frac{2R}n)^2}\right)=\exp\left(-\frac{n\e^2}{32R^2}\right).$$

\subsection{Proof of Lemma \ref{lem:TnTn*}}
For $x\in\R^d$,\begin{align*} 
|\T_n^\star(x)-\T_n(x)|&=\left|\frac{\sum_{i=1}^nY_iK_\alpha(Z_i-z)}{n\E [K_\alpha(Z-z)]}-\frac{\sum_{i=1}^nY_iK_\alpha(Z_i-z)}{\sum_{i=1}^nK_\alpha(Z_i-z)}\right|
\\&=\left|\sum_{i=1}^nY_iK_\alpha(Z_i-z)\right|\left|\frac{1}{n\E [K_\alpha(Z-z)]}-\frac{1}{\sum_{i=1}^nK_\alpha(Z_i-z)}\right|\\&\leq
\left|\sum_{i=1}^nK_\alpha(Z_i-z)\right|\left|\frac{1}{n\E [K_\alpha(Z-z)]}-\frac{1}{\sum_{i=1}^nK_\alpha(Z_i-z)}\right|\\&
=\left|\frac{\sum_{i=1}^nK_\alpha(Z_i-z)}{n\E [K_\alpha(Z-z)]}-1\right|.
\end{align*}
So, by Lemma \ref{lem:etaTn*}, applied in the particular case where $Y$ is constant, equal to 1, we obtain for every $\e>0$, $$\P\left(\int |\T_n^\star(x)-\T_n(x)|\mu(dx)>\e/2\right)\leq\exp\left(-\frac{n\e^2}{8R^2}\right).$$

\subsection{Proof of the main theorems}

Finally, to prove Theorem \ref{theo:reg}, it suffices to use Equation \eqref{eq:dec1} and to combine Lemma \ref{lem:etaTn*} and Lemma \ref{lem:TnTn*}. For every $\e>0$, we obtain, as soon as  $n$ is large enough, 
\begin{align*}
\P\left(\int |\eta(x)-\T_n(x)|\mu(dx)>\e\right)&\leq \P\left(\int|\eta(x)-\T_n^\star(x)|\mu(dx)>\e/2\right)+\P\left(\int |\T_n^\star(x)-\T_n(x)|\mu(dx)>\e/2\right)\\&\leq\exp\left(-\frac{n\e^2}{32R^2}\right)
+\exp\left(-\frac{n\e^2}{8R^2}\right)\\&\leq 2\exp\left(-\frac{n\e^2}{32R^2}\right).
\end{align*}
For Theorem \ref{theo:class}, Equation \eqref{eq:declass} and Lemma \ref{lem:etaTn*} show that, for every $\e>0$, if $n$ is large enough, 
\begin{multline*}
\P(L_n-L^\star>\e)\\\leq \P\left(\int\left|1-\eta(x)-\frac{\sum_{i=1}^n  K_\alpha(Z_i-z)}{n\E  [K_\alpha(Z-z)]}+\T_n^\star(x)\right|\mu(dx)>\e/2\right)
+\P\left(\int|\eta(x)-\T_n^\star(x)|\mu(dx)>\e/2\right)\\\leq 2\exp\left(-\frac{n\e^2}{32R^2}\right).
\end{multline*} 

\section{Annex}

\begin{itemize}
	\item The ``gauss'' example is a mixture of two Gaussian distributions ${\cal N}\left(\left(\begin{smallmatrix}0\\2\end{smallmatrix}\right), \left(\begin{smallmatrix}1&-0.5\\ -0.5& 1\end{smallmatrix}\right)\right)$ and  ${\cal N}\left(\left(\begin{smallmatrix}-1\\2\end{smallmatrix}\right),\left(\begin{smallmatrix}1&0.5\\ 0.5& 1\end{smallmatrix}\right)\right)$.

\item The ``comete'' example is a mixture of two Gaussian distributions ${\cal N}\left(\left(\begin{smallmatrix}0\\2\end{smallmatrix}\right), \left(\begin{smallmatrix}3&9/4\\9/4&15\end{smallmatrix}\right)\right)$ and  ${\cal N}\left(\left(\begin{smallmatrix}0\\2\end{smallmatrix}\right), \left(\begin{smallmatrix}1& 0\\0 & 1\end{smallmatrix}\right)\right)$.

\item The ``nuclear'' example is a mixture of a uniform distribution ${\cal U} (\mathcal D^1)$ on the unit disk 
and a  Gaussian distribution  ${\cal N}\left(\left(\begin{smallmatrix}0.5\\0.5\end{smallmatrix}\right), \left(\begin{smallmatrix}0.1&0\\0&0.1\end{smallmatrix}\right)\right)$.

\item The ``spot'' example is a mixture of a uniform ${\cal U}  ([-1,1]^2)$
and 5  Gaussian distributions  arbitrary chosen on the unit square.


\end{itemize}

\section{Acknowledgement}
This research has been partially supported by French Research National Agency (ANR) as part of the project FOREWER (reference:  ANR-14-CE05-0028).

\bibliographystyle{plainnat}
\bibliography{biblio-mixcobra}

\end{document}